\documentclass{article}
\PassOptionsToPackage{numbers,compress}{natbib}
\usepackage[main,final]{neurips_2026}
\usepackage[utf8]{inputenc}
\usepackage[T1]{fontenc}
\usepackage{float}
\usepackage{hyperref}
\usepackage{url}
\usepackage{booktabs}
\usepackage{amsfonts}
\usepackage{amssymb}
\usepackage{amsmath}
\usepackage{amsthm}
\usepackage{nicefrac}
\usepackage{microtype}
\usepackage{xcolor}
\usepackage{graphicx}
\usepackage{multirow}
\usepackage{enumitem}

\newtheorem{theorem}{Theorem}
\newtheorem{proposition}{Proposition}

\newtheorem{lemma}{Lemma}

\makeatletter
\renewenvironment{abstract}%
{%
  \vskip 0.075in%
  \centerline{\large\bf Abstract}%
  \vspace{0.5ex}%
  \list{}{%
    \leftmargin=1.6em%
    \rightmargin=1.6em%
  }%
  \item\relax
}
{%
  \endlist%
  \vskip 1ex%
}
\makeatother

\title{Confidence-Calibrated Inference Expansion for Evaluator-Guided Test-Time Reasoning}

\author{%
  Weida Liang \\
  School of Computing \\
  National University of Singapore \\
  \texttt{weida\_liang@u.nus.edu}
  \And
  Kenji Kawaguchi \\
  School of Computing \\
  National University of Singapore \\
  \texttt{kenji@comp.nus.edu.sg}
}

\begin{document}
\maketitle

\begin{abstract}
Evaluator-guided reasoning presents a local control problem at every step: trust the current answer, return a fallback, or spend more inference. We formulate this choice as a compute-aware three-action controller, scoring terminal actions by predicted answer success and expansion by a rollout-based continuation-success surrogate, with compute cost priced explicitly. A local action-gap analysis characterizes when score errors can change a decision, while an expected regret bound accounts for errors along the realized expansion path. Guided by this analysis, we learn and calibrate action-specific confidence estimates, with local uncertainty penalties for ambiguous decisions. On held-out width expansion and revision tasks, the controller improves observed utility over calibration-selected fixed policies while reducing normalized online compute by about 30\% and 16\%, respectively. Across four Qwen3-8B seeds, compute is lower than fitted Q in both settings, while utility differences are small and mixed. The results frame evaluator-guided reasoning as resource allocation under uncertainty, rather than a uniformly stronger reasoner.
\end{abstract}
\section{Introduction}

Adaptive test-time reasoning increasingly relies on evaluator-guided continuation: sampling more
chains, reranking with a verifier, or revising a draft under judge feedback
\citep{
wei2022chain,
wang2022self,
cobbe2021training,
uesato2022solving,
lightman2023let,
yao2023tree,
madaan2023self,
shinn2023reflexion,
brown2024large,
snell2024scaling,
lifshitz2025multi,
setlur2024rewarding}.
These strategies can improve average performance as more inference compute is spent
\citep{brown2024large,snell2024scaling,guo2025deepseek}, but deployment turns inference allocation
into a local control problem: given a current proposal and evaluator evidence, is the next unit of
inference worth its cost for this instance?

This decision is broader than standard answer selection. At an intermediate inference state, the system
may stop with the evaluator-preferred answer, stop conservatively with a fallback answer, or buy one
more unit of inference and decide again. We study this as a three-action controller over
\[
\mathcal A(s)=\{\texttt{trust},\texttt{fallback},\texttt{expand}\}.
\]
Here, \texttt{trust} emits the current evaluator-preferred answer, \texttt{fallback} emits a
conservative answer that avoids further evaluator-guided continuation, and \texttt{expand} buys one
more inference step before the controller acts again. This abstraction covers both width expansion
(sample more candidates) and revision expansion (run another critique-and-revision round). It is
therefore broader than selective prediction and reject-option classification
\citep{chow2003optimum,hellman2007nearest,bartlett2008classification,el2010foundations,geifman2017selective,cortes2016learning}:
low confidence need not imply rejection, because additional inference may itself be the right action.

The control problem is not answered by raw evaluator scores alone. Verifiers, reward models, and
process reward models are useful for ranking or guiding reasoning
\citep{cobbe2021training,uesato2022solving,lightman2023let,setlur2024rewarding},
but a high verifier score does not by itself guarantee that stopping is best, and a low judge score does
not automatically make fallback the right choice. For terminal actions, the relevant quantity is
\emph{action-specific confidence}: the probability that the emitted answer succeeds. Expansion instead
has a Bellman continuation value that includes downstream actions and their costs. Our controller
estimates terminal success and a rollout-based continuation-success surrogate for expansion, calibrates
these probabilities \citep{platt1999probabilistic,zadrozny2002transforming,guo2017calibration,kumar2019verified},
subtracts the immediate compute cost, and optionally penalizes actions with high local uncertainty.

We formalize this decision problem with compute-adjusted values: terminal actions receive answer
success probability minus cost, while expansion receives its Bellman continuation value minus the
immediate cost. This framing yields two controller-level results. Local action selection is stable when
score error is smaller than the compute-adjusted action gap, and regret accumulates along states visited
during expansion. The expansion head's success-based target approximates the continuation value but
does not include downstream compute costs.

We evaluate verifier-guided width expansion on MATH~\citep{hendrycks2021measuring} and AIME25
\citep{maa_aime_2025_misc}, and judge-guided revision on a deterministic-answer General Reasoning
benchmark filtered from OpenThoughts-114k
\citep{guha2025openthoughtsdatarecipesreasoning}. Scores use reference matching independent of
the revision judge. Across Qwen3-4B, Qwen3-8B~\citep{yang2025qwen3}, and
DeepSeek-R1-Distill-Qwen-14B~\citep{guo2025deepseek}, the controller reaches lower-compute
operating points than strong fixed policies across the two reasoning settings. We also examine target
sensitivity, action behavior, and compute-price retuning.

\paragraph{Contributions.}
\begin{itemize}[leftmargin=1.5em]
    \item We formulate evaluator-guided reasoning as a compute-aware controller over \texttt{trust},
    \texttt{fallback}, and \texttt{expand}, covering width and revision expansion.
    \item We analyze how action-confidence error interacts with the compute-adjusted action gap and
    how regret accumulates along the realized expansion path.
    \item We develop a lightweight controller with one learned confidence head per action, post-hoc
    calibration, and an optional uncertainty penalty estimated from held-out residuals.
    \item We evaluate compute allocation, controller components, target sensitivity, and action behavior
    across two reasoning settings.
\end{itemize}
\section{Related Work}

\subsection{Adaptive inference and compute allocation}

Test-time reasoning methods use sampling, self-consistency, verifier-guided reranking, tree search, and
self-refinement to improve outputs with additional inference compute
\citep{wei2022chain,kojima2022large,wang2022self,cobbe2021training,yao2023tree,madaan2023self,shinn2023reflexion,brown2024large,snell2024scaling}. Recent scaling studies show that returns depend on task difficulty, evaluator quality, and allocation strategy
\citep{setlur2025scaling}. Most methods choose a budget before inference or study how to use a budget
once fixed. We instead ask at each intermediate state whether to trust, fall back, or continue.

Adaptive routers and value-based allocation methods also optimize quality against cost, including
pre-inference routers, Lagrangian allocation, adaptive best-of-\(N\), and reward- or length-aware
self-evaluation \citep{zhai2026adaptive,manvi2025zero,manvi2024adaptive,ding2025best}. Re-FORC
predicts future reward as a function of additional reasoning tokens, enabling adaptive stopping and
compute allocation, while Pandora-style search models sequential acquisition with explicit inspection
costs \citep{zabounidis2026reforc,weitzman1979optimal}. Unlike a
budget choice alone, our action set includes two terminal answers, trust and fallback, as well as
continuation. The controller predicts terminal-answer success and uses a rollout-based
continuation-success surrogate for expansion, while pricing compute separately.

\subsection{Evaluators, fallback, and calibration}

Verifiers, reward models, process rewards, and judges guide candidate selection, critique, and search
\citep{uesato2022solving,lightman2023let,setlur2025scaling,lifshitz2025multi}. In revision, however,
self-correction is not monotone and can corrupt correct drafts \citep{madaan2023self,shinn2023reflexion,huang2023large}. We therefore treat evaluator feedback as an input to a stop-or-continue decision. Fallback
is not abstention in selective prediction \citep{chow2003optimum,bartlett2008classification,el2010foundations,geifman2017selective,cortes2016learning}: it emits a conservative answer, while expansion can acquire further evidence.

Post-hoc calibration and conformal risk control provide tools for relating confidence to empirical
outcomes \citep{platt1999probabilistic,zadrozny2002transforming,guo2017calibration,jiang2018trust,angelopoulos2022conformal}. Because trust, fallback, and expand induce different success events, a single
correctness score can obscure decision-relevant differences. We estimate and calibrate one confidence
per action, then use local residual uncertainty to temper choices near small compute-adjusted action
gaps.
\section{Problem Setup}

\subsection{State, evidence, and actions}

At inference time, the system is in a state
\[
s=(x,\xi,\zeta,b),
\]
where \(x\) is the input, \(\xi\) is the current reasoning state, \(\zeta\) is evaluator-side evidence,
and \(b\) is the compute already spent. From this state, the controller observes a current proposal
\(y_{\mathrm{cur}}(s)\), a conservative fallback answer \(y_0(s)\), and features \(\phi(s)\) constructed
only from inference-time evidence.

The available actions are
\[
\mathcal A(s)=\{\texttt{trust},\texttt{fallback}\}\cup\{\texttt{expand}: b<b_{\max}\}.
\]
The terminal actions emit
\[
\texttt{trust}: y_{\mathrm{cur}}(s),
\qquad
\texttt{fallback}: y_0(s),
\]
while \texttt{expand} buys one more inference step and transitions to a richer state \(s^+\).

\subsection{Two expansion mechanisms}

We study two finite-horizon instantiations of \texttt{expand}. In \emph{width expansion}, \(\xi\) is
a candidate pool, \(y_{\mathrm{cur}}\) is the verifier-best candidate, \(y_0\) is majority vote over
the current pool, and expansion moves to the next width on a fixed ladder. In \emph{revision
expansion}, \(\xi\) is the current draft plus revision history, \(y_{\mathrm{cur}}\) is the latest draft,
\(y_0\) is a rollback answer, and expansion executes one more judge-guided critique-and-revision
round. Thus, fallback is a controller role instantiated differently across settings: majority vote in
width expansion and rollback in revision expansion. In both cases, the budget is finite, so continuation
values are well defined by backward recursion.

\subsection{Action-specific confidence and compute-aware value}

For terminal actions, the ideal compute-adjusted value is
\[
J^\star(s,u)=p^\star(s,u)-\lambda c(s,u),
\qquad
u\in\{\texttt{trust},\texttt{fallback}\},
\]
where \(p^\star(s,u)\) is the success probability of the emitted answer and \(\lambda\) prices test-time
computation. This probability is distinct from raw verifier or judge confidence. For expansion,
\[
J^\star(s,\texttt{expand})
=
-\lambda c(s,\texttt{expand})
+
\mathbb E\left[V^\star(s^+)\mid s,\texttt{expand}\right],
\qquad
V^\star(s)=\max_{u\in\mathcal A(s)}J^\star(s,u).
\]
The ideal value of \texttt{expand} is the Bellman continuation value, which includes the costs and
outcomes of subsequent actions.

At deployment, \(\widehat p(s,u)\) estimates terminal-answer success for \texttt{trust} and
\texttt{fallback}, and a rollout-defined continuation-success probability for \texttt{expand}. The
expansion estimate is a success-based surrogate for, rather than a direct estimate of, the Bellman
continuation value: it does not include downstream compute costs. The controller forms
\[
\widehat J(s,u)=\widehat p(s,u)-\lambda c(s,u).
\]
The simplest online policy chooses \(\arg\max_u \widehat J(s,u)\). Section~\ref{sec:controller}
specifies how we learn and calibrate \(\widehat p(s,u)\), and how we incorporate an uncertainty
penalty near brittle action boundaries.
\section{Theoretical Analysis}
\label{sec:analysis}

We analyze when the online controller agrees with the ideal finite-horizon controller and how local
confidence errors propagate along an expansion trajectory. The main text states the controller-level
results that motivate the method. Appendix~\ref{app:proofs} gives full proofs and supplementary
results on certified decisions, over-revision, width fragility, and scalar-confidence limitations.

\subsection{Local action consistency}

Let \(\widetilde J(s,u)\) be the score used by the deployed policy to select an available action.
For the uncertainty-aware controller,
\(\widetilde J(s,u)=\widehat p(s,u)-\lambda c(s,u)-\beta\widehat\epsilon_p(s,u)\).
Define the local score error
\[
\delta_J(s)=\max_{u\in\mathcal A(s)}\left|\widetilde J(s,u)-J^\star(s,u)\right|
\]
and the compute-adjusted action gap
\[
\Gamma^\star(s)=J^\star(s,\pi^\star(s))-\max_{u\neq \pi^\star(s)}J^\star(s,u).
\]
The expansion score is a success-based surrogate rather than a direct estimate of the Bellman action
value. Accordingly, \(\delta_J(s)\) includes continuation-target mismatch and omitted future-cost
effects, in addition to confidence-estimation error.
When \(\Gamma^\star(s)\) is small, even a calibrated controller may flip actions because multiple
decisions have similar value.

\begin{lemma}[Local action consistency]
\label{lem:local_action_consistency}
If \(2\delta_J(s)<\Gamma^\star(s)\), then \(\widehat\pi(s)=\pi^\star(s)\).
\end{lemma}

The lemma identifies the relevant reliability target: the action-selection score after compute pricing
and uncertainty adjustment, rather than a raw evaluator score. Verifier or judge margins enter only as
features in \(\phi(s)\), not as surrogates for \(\Gamma^\star(s)\).

\subsection{Trajectory-level regret}

For any policy \(\pi\), let \(V^\pi(s)\) denote its finite-horizon value and define its regret by
\[
R_\pi(s)=V^\star(s)-V^\pi(s).
\]

\begin{theorem}[Finite-horizon confidence regret]
\label{thm:finite_horizon_regret}
For a policy \(\widehat\pi\) that selects an action maximizing its deployed score \(\widetilde J\),
\[
R_{\widehat\pi}(s)
\le
2\delta_J(s)
+
\mathbf 1\{\widehat\pi(s)=\texttt{expand}\}
\mathbb E\left[R_{\widehat\pi}(s^+)\mid s,\texttt{expand}\right].
\]
For an initial state \(s_0\), let \(S_0=s_0,S_1,\ldots,S_T\) be the random states visited under
\(\widehat\pi\), with \(T+1\leq H\). Then
\[
R_{\widehat\pi}(s_0)
\leq 2\,\mathbb E_{\widehat\pi}\!\left[\sum_{t=0}^{T}\delta_J(S_t)\mid S_0=s_0\right].
\]
\end{theorem}

The bound controls expected cumulative score error along states reached by the policy.

\subsection{Visited-state coverage and practical control}

Assume an action-specific uncertainty radius \(\widehat\epsilon_J(s,u)\) covers error in the deployed
score on the policy's reachable states. If the unpenalized score error is covered by
\(\widehat\epsilon_p\), the uncertainty penalty gives a radius of
\((1+\beta)\widehat\epsilon_p\) for the deployed score. For \texttt{expand}, however, the
calibration residual is measured against a binary continuation-success target and does not by itself
establish coverage of the Bellman score error. Writing
\(\widehat\epsilon_{\max}(s)=\max_{u\in\mathcal A(s)}\widehat\epsilon_J(s,u)\), we obtain the
following conditional bound.

\begin{theorem}[Trajectory regret under visited-state coverage]
\label{thm:high_probability_regret}
Let \(\mathcal R_{\widehat\pi}(s_0)\) be the set of states reachable with positive probability from
\(s_0\) under \(\widehat\pi\) before termination. Suppose a joint event \(\mathcal E\) has probability
at least \(1-\alpha\) and, on \(\mathcal E\),
\[
\left|\widetilde J(s,u)-J^\star(s,u)\right|
\le \widehat\epsilon_J(s,u)
\qquad
\forall s\in\mathcal R_{\widehat\pi}(s_0),\ \forall u\in\mathcal A(s).
\]
Then, on \(\mathcal E\),
\[
R_{\widehat\pi}(s_0)
\leq 2\,\mathbb E_{\widehat\pi}\!\left[\sum_{t=0}^{T}\widehat\epsilon_{\max}(S_t)
\mid S_0=s_0\right].
\]
The bound therefore holds with probability at least \(1-\alpha\).
\end{theorem}

The local consistency condition depends on the compute-adjusted action gap. Revisions can repair or
corrupt a draft, and wider candidate pools can improve coverage while increasing sensitivity to
evaluator noise.
\section{Main Method}
\label{sec:controller}

\subsection{Action-specific confidence heads}

We use a lightweight controller so that decision overhead is negligible relative to generation,
verification, judging, and revision. For each state \(s\), the feature vector \(\phi(s)\) contains only
inference-time quantities: evaluator score, score margin, uncertainty or disagreement summary,
agreement or rollback support, recent progress, expansion level, and normalized budget.
Appendix~\ref{app:features} lists the width-specific and revision-specific feature definitions.

We fit one sigmoid confidence head per action,
\[
\widehat p_{\mathrm{raw}}(s,u)=\sigma(w_u^\top \phi(s)),
\qquad
u\in\{\texttt{trust},\texttt{fallback},\texttt{expand}\}.
\]
Thus, online decisions are made by three small action-specific heads over observable features, rather
than by an additional search procedure.

\subsection{Offline targets and calibration}

Offline full-ladder rollouts provide controller-aligned supervision. For \texttt{trust} and
\texttt{fallback}, the target is whether the emitted answer succeeds. For \texttt{expand}, the target is
whether the rollout-selected empirical continuation after one expansion step eventually succeeds.
This binary target is a success-based surrogate rather than a Bellman return; the controller subtracts
the immediate expansion cost separately.

We train the raw heads with binary cross-entropy on a fit split and calibrate each action head on a
disjoint calibration split,
\[
\widehat p(s,u)=\mathrm{Cal}_u\bigl(\widehat p_{\mathrm{raw}}(s,u)\bigr).
\]
The deployed surrogate value is
\[
\widehat Q(s,u)=\widehat J(s,u)=\widehat p(s,u)-\lambda c(s,u),
\]
which is the controller-level score used for action selection. Appendix~\ref{app:splits_targets}
compares oracle-aligned, one-step, and policy-induced \texttt{expand} targets. Full rollout
construction, splits, training, and calibration details appear in Appendix~\ref{app:splits_targets}--
\ref{app:controller_training} and Appendix~\ref{app:calibration}.

\subsection{Uncertainty-aware deployment}

To discourage brittle decisions near action boundaries, we estimate an action-specific local uncertainty
radius \(\widehat\epsilon_p(s,u)\) from residuals on the disjoint calibration split, using coarse buckets over
expansion level, margin, uncertainty, and budget. Appendix~\ref{app:calibration} describes the bucket
construction and sparse-bucket handling.

The deployed policies are
\[
\pi_{\mathrm{raw}}(s)=\arg\max_{u\in\mathcal A(s)}
\left[\widehat p_{\mathrm{raw}}(s,u)-\lambda c(s,u)\right],
\]
\[
\pi_{\mathrm{conf}}(s)=\arg\max_{u\in\mathcal A(s)}
\left[\widehat p(s,u)-\lambda c(s,u)\right],
\]
and our main controller
\[
\pi_{\mathrm{ours}}(s)=\arg\max_{u\in\mathcal A(s)}
\left[\widehat p(s,u)-\lambda c(s,u)-\beta\widehat\epsilon_p(s,u)\right].
\]
The scalar \(\beta\) is tuned on the calibration split. Setting \(\beta=0\) recovers the calibrated
controller. At deployment, the controller uses only online features, calibrated action confidences,
explicit compute cost, and a bucket lookup for \(\widehat\epsilon_p\). Offline ladders are required
only for controller fitting and evaluation, not for online execution.
\section{Experiments}

We evaluate Qwen3-4B, Qwen3-8B~\citep{yang2025qwen3}, and DeepSeek-R1-Distill-Qwen-14B
~\citep{guo2025deepseek} without fine-tuning. In verifier-guided width expansion, the system starts
with four MATH or AIME25 candidates and may expand through widths 8, 16, 24, 32, and 48. The
verifier-best answer is the proposal and majority vote is the fallback. In judge-guided revision, a
first draft may undergo up to four critique-and-revision rounds on a balanced 600-example
OpenThoughts-derived benchmark. Final success uses exact or deterministic reference matching and is
scored independently of the revision judge. Dataset construction, prompts, filtering, and feature
definitions are described in Appendix~\ref{app:benchmark_details}--\ref{app:features}.

Offline ladders are divided into fit, calibration, and held-out test splits. The deployment objective is
\[
U=S-\lambda C,\qquad \lambda=0.02,
\]
where \(C\) is total online prompt and generation compute, normalized by the corresponding max-budget
baseline. Utility is recomputed from unrounded success and compute before aggregation. Offline ladder
materialization is required for fitting and evaluation but is excluded from per-query cost. We report
held-out success, compute-adjusted utility, and normalized online compute. The offline ladder oracle is
used to construct controller-training targets.

Baselines include fixed and maximum budgets, threshold and rollback heuristics, raw and calibrated
confidence, budget-conditioned value models, and a fitted-Q controller.
Fixed policies and controller hyperparameters are selected on calibration data under the same objective.
Appendix~\ref{app:baseline_tuning} and Appendix~\ref{app:additional_results} give
full tuning, uncertainty, ablation, action, and cost analyses.
\section{Results}
\label{sec:results}

\paragraph{Overview.}
At the nominal compute price, the controller reaches higher utility with lower online compute than
the calibration-selected fixed policy in both width expansion and revision. Four independent
Qwen3-8B seeds show lower compute than fitted Q in every run, while utility differences favor the
controller in two of four width seeds and three of four revision seeds.

\subsection{Scenario~I: verifier-guided width expansion}

Table~\ref{tab:scenario1_main} summarizes results on MATH and AIME25. Confidence plus uncertainty
reaches utility 0.715 at compute 0.547, matching majority vote at \(N=48\) while using 0.453 less
normalized compute. Relative to the calibration-selected fixed-width policy at \(N=32\), it improves
utility from 0.707 to 0.715 and reduces compute from 0.781 to 0.547. Calibrated confidence alone
reaches utility 0.714 at compute 0.586.

\begin{table}[t]
\centering
\small
\setlength{\tabcolsep}{3pt}
\caption{Scenario~I on MATH and AIME25. Best fixed width is majority vote at \(N=32\), selected on calibration data. Naive maximum width is verifier selection at \(N=48\) and sets normalized compute to 1.0.}
\label{tab:scenario1_main}
\begin{tabular*}{\columnwidth}{@{\extracolsep{\fill}}lcccc}
\toprule
Method & MATH Acc. & AIME25 Acc. & Mean Utility & Compute \\
\midrule
Trust at \(N=4\) & 0.742 & 0.469 & 0.601 & \textbf{0.240} \\
Best fixed: majority, \(N=32\) & 0.821 & 0.625 & 0.707 & 0.781 \\
Max width: verifier, \(N=48\) & 0.810 & 0.642 & 0.706 & 1.000 \\
Majority vote, \(N=48\) & 0.824 & 0.646 & 0.715 & 1.000 \\
Margin heuristic & 0.807 & 0.594 & 0.688 & 0.634 \\
Uncalibrated confidence & 0.819 & 0.608 & 0.701 & 0.611 \\
Calibrated confidence & \textbf{0.826} & 0.625 & 0.714 & 0.586 \\
Confidence + uncertainty (ours) & 0.824 & 0.628 & \textbf{0.715} & 0.547 \\
\bottomrule
\end{tabular*}
\end{table}

The fixed-width sweep in Table~\ref{tab:fixed_width_sweep_app} compares verifier selection and
majority vote across the evaluated widths. Majority vote improves with width, while the adaptive
controller matches its width-48 utility at substantially lower compute.

\begin{table}[t]
\centering
\small
\setlength{\tabcolsep}{3pt}
\caption{Held-out fixed-width sweep at \(\lambda=0.02\). Utility is averaged across MATH and AIME25; compute is normalized to verifier trust at \(N=48\).}
\label{tab:fixed_width_sweep_app}
\begin{tabular*}{\columnwidth}{@{\extracolsep{\fill}}lcc}
\toprule
Width policy & Mean utility & Compute \\
\midrule
Trust at \(N=4\) & 0.601 & 0.240 \\
Verifier trust at \(N=8\) & 0.652 & 0.389 \\
Verifier trust at \(N=16\) & 0.687 & 0.579 \\
Verifier trust at \(N=24\) & 0.694 & 0.721 \\
Verifier trust at \(N=32\) & 0.693 & 0.781 \\
Verifier trust at \(N=48\) & 0.706 & 1.000 \\
Majority vote at \(N=16\) & 0.687 & 0.579 \\
Majority vote at \(N=32\) & 0.707 & 0.781 \\
Majority vote at \(N=48\) & 0.715 & 1.000 \\
Confidence + uncertainty (ours) & 0.715 & 0.547 \\
\bottomrule
\end{tabular*}
\end{table}

\subsection{Scenario~II: judge-guided revision expansion}

Table~\ref{tab:scenario2_main} reports judge-guided revision, where \texttt{expand} runs one
additional critique-and-revision round. The controller reaches utility 0.673 at compute 0.529,
compared with utility 0.668 at compute 0.632 for the calibration-selected fixed-depth policy. The
maximum-round policy reaches success 0.688 at full budget.

\begin{table}[t]
\centering
\small
\setlength{\tabcolsep}{3pt}
\caption{Scenario~II on General Reasoning at \(\lambda=0.02\). Fixed-depth baselines are selected on calibration data under the same objective as the adaptive controller.}
\label{tab:scenario2_main}
\begin{tabular*}{\columnwidth}{@{\extracolsep{\fill}}lccc}
\toprule
Method & Success & Utility & Compute \\
\midrule
First-draft trust & 0.603 & 0.598 & \textbf{0.240} \\
Best fixed-depth revision & 0.681 & 0.668 & 0.632 \\
Naive max-round revision & \textbf{0.688} & 0.668 & 1.000 \\
Rollback heuristic & 0.668 & 0.657 & 0.557 \\
Uncalibrated confidence & 0.667 & 0.655 & 0.606 \\
Calibrated confidence & 0.680 & 0.668 & 0.579 \\
Confidence + uncertainty (ours) & 0.684 & \textbf{0.673} & 0.529 \\
\bottomrule
\end{tabular*}
\end{table}

\subsection{Calibration, coverage, and seed variation}

Calibration lowers ECE for every action head. For \texttt{trust}/\texttt{fallback}/\texttt{expand},
ECE changes from 0.061/0.066/0.084 to 0.024/0.029/0.041 in width and from
0.069/0.073/0.096 to 0.031/0.035/0.052 in revision. Table~\ref{tab:coverage_target_sensitivity}
shows the operating points across nominal coverage targets; at the 90\% target, held-out expand
coverage is 0.882 in width and 0.861 in revision.
The 90\% target is the prespecified main setting; at 95\%, held-out utility is slightly higher and
compute is lower in both width (0.716/0.522 vs. 0.715/0.547) and revision
(0.675/0.503 vs. 0.673/0.529), reported as utility/compute.

\begin{table}[t]
\centering
\small
\setlength{\tabcolsep}{3pt}
\caption{Sensitivity to nominal coverage target at \(\lambda=0.02\). Expand coverage is measured on held-out states.}
\label{tab:coverage_target_sensitivity}
\begin{tabular*}{\columnwidth}{@{\extracolsep{\fill}}lcccc}
\toprule
Scenario & Target coverage & Utility & Compute & Expand coverage \\
\midrule
Width & 70\% & 0.713 & 0.584 & 0.714 \\
Width & 80\% & 0.715 & 0.566 & 0.803 \\
Width & 90\% & 0.715 & 0.547 & 0.882 \\
Width & 95\% & 0.716 & 0.522 & 0.934 \\
Revision & 70\% & 0.672 & 0.570 & 0.706 \\
Revision & 80\% & 0.671 & 0.549 & 0.792 \\
Revision & 90\% & 0.673 & 0.529 & 0.861 \\
Revision & 95\% & 0.675 & 0.503 & 0.925 \\
\bottomrule
\end{tabular*}
\end{table}

Four Qwen3-8B generation seeds use one fixed problem split, with all methods refit and recalibrated
within each seed. Table~\ref{tab:seed_fitted_q} reports paired Student-$t$ intervals over the four
seeds. Utility intervals include zero in both settings, while compute intervals are below zero.

\begin{table}[t]
\centering
\small
\setlength{\tabcolsep}{3pt}
\caption{Four-seed paired comparison with fitted Q on a fixed Qwen3-8B problem split. All methods are refit and recalibrated within each generation seed. Paired \(\Delta U\) and \(\Delta C\) are reported as Student-$t$ 95\% intervals with three degrees of freedom.}
\label{tab:seed_fitted_q}
\begin{tabular*}{\columnwidth}{@{\extracolsep{\fill}}lccc}
\toprule
Scenario & \shortstack{Paired $\Delta U$\\95\% CI} & \shortstack{Paired $\Delta C$\\95\% CI} & Seed direction \\
\midrule
Width & $-0.001$ [$-0.006,0.004$] & $-0.037$ [$-0.055,-0.020$] & $U$ 2/4, $C$ 4/4 \\
Revision & $+0.002$ [$-0.003,0.007$] & $-0.027$ [$-0.040,-0.014$] & $U$ 3/4, $C$ 4/4 \\
\bottomrule
\end{tabular*}
\end{table}

The seed-level differences in Appendix~\ref{app:additional_results} show that compute is lower in
every seed, while utility differences vary in direction.

\subsection{Controller diagnostics}

Evaluator margin correlates with the empirical controller action gap at 0.426 in width and 0.368 in
revision. Among action changes, 62\% in width and 59\% in revision occur in small-gap states
(Table~\ref{tab:action_diagnostics_app}).

\section{Analysis and Ablations}
\label{sec:analysis_ablations}

\subsection{Scale, difficulty, and controller components}

Across backbones, mean width compute falls from 0.644 for Qwen3-4B to 0.434 for
DeepSeek-R1-Distill-Qwen-14B. Revision compute falls from 0.606 to 0.446, while mean rounds fall
from 1.680 to 0.940. Width allocation also tracks difficulty: MATH initial expansion rises from 0.18
at Level 1 to 0.57 at Level 5, and mean expansions rise from 0.62 to 1.61. Per-backbone results are
reported in Table~\ref{tab:backbone_results_app}.

With shared predictive weights and per-action calibration, utility/compute is 0.710/0.595 in width
and 0.662/0.585 in revision. Separate heads with per-action calibration reach 0.714/0.586 and
0.668/0.579; adding local uncertainty yields 0.715/0.547 and 0.673/0.529. The full component
comparison is in Table~\ref{tab:shared_head_ablation_app}; Table~\ref{tab:component_summary}
isolates the main steps.

\begin{table}[t]
\centering
\small
\setlength{\tabcolsep}{3pt}
\caption{Controller-component ablation at \(\lambda=0.02\). Cells report utility/compute.}
\label{tab:component_summary}
\begin{tabular*}{\columnwidth}{@{\extracolsep{\fill}}lcc}
\toprule
Controller & Width \(U/C\) & Revision \(U/C\) \\
\midrule
Shared weights, per-action calibration & 0.710 / 0.595 & 0.662 / 0.585 \\
Separate heads, per-action calibration & 0.714 / 0.586 & 0.668 / 0.579 \\
Separate heads + local uncertainty & 0.715 / 0.547 & 0.673 / 0.529 \\
\bottomrule
\end{tabular*}
\end{table}

Separate per-action calibration adds 0.004 utility in width and 0.006 in revision relative to shared
predictive weights. Local uncertainty adds a further 0.001 and 0.005 while reducing compute by
0.039 and 0.050. Retaining fallback adds 0.006 utility in width and 0.010 in revision over the
trust/expand-only controller (Table~\ref{tab:fallback_ablation_app}).

Policy-induced continuation targets reach utility 0.714 in width and 0.669 in revision, compared
with 0.715 and 0.673 for oracle-aligned targets. One-step targets reach 0.709 and 0.665. Joint
fit-and-calibration corruption from 10\% to 30\% lowers utility from 0.709 to 0.689 in width and
from 0.664 to 0.642 in revision. Tables~\ref{tab:expand_target_ablation_app} and
\ref{tab:target_corruption_app} report the full results.

\subsection{Action behavior and revision dynamics}

\begin{figure}[t]
\centering
\includegraphics[width=0.96\textwidth]{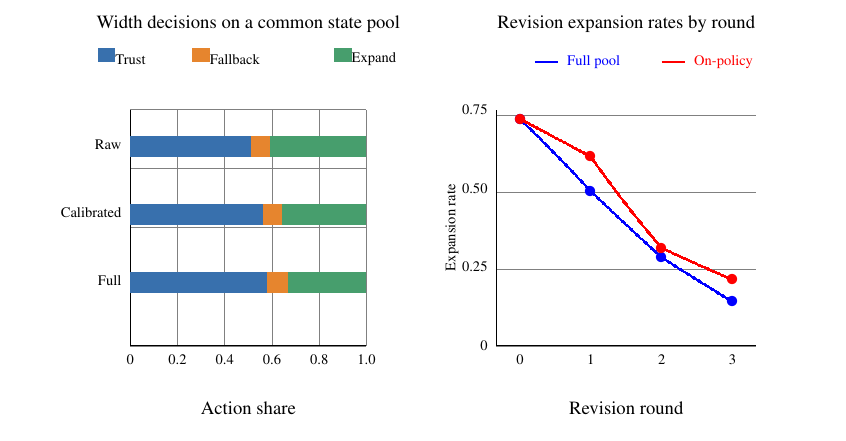}
\caption{Counterfactual width action shares on a common state pool (left) and revision expansion rates by round (right). The revision panel distinguishes full-pool rates from on-policy conditional hazards.}
\label{fig:action_selection}
\end{figure}

Calibration most often changes \texttt{expand} to \texttt{trust} in width and \texttt{expand} to
\texttt{fallback} in revision. Most action flips occur in small-gap states. Full-cohort revision
success rises from 0.603 at zero rounds to 0.681 at two rounds, falls to 0.673 at three, and reaches
0.688 at four. Action shares and transition
rates are reported in Tables~\ref{tab:action_diagnostics_app} and
\ref{tab:revision_round_outcomes_app}. Figure~\ref{fig:action_selection} shows the resulting
width action shares and revision expansion rates by depth.

\subsection{Compute price and alternative value models}

\begin{figure}[t]
\centering
\includegraphics[width=0.82\textwidth]{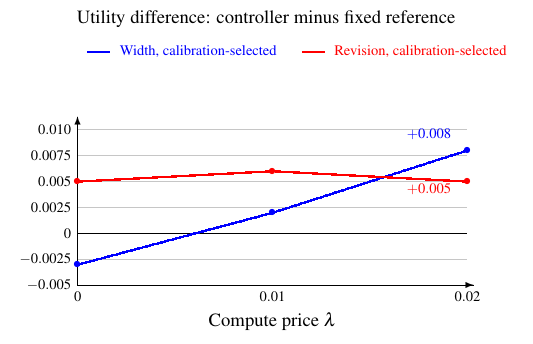}
\caption{Retuned utility differences for compute prices \(\lambda\in\{0,0.01,0.02\}\), relative to calibration-selected fixed policies.}
\label{fig:lambda_retuned}
\end{figure}

Figure~\ref{fig:lambda_retuned} reports retuned utility differences against calibration-selected fixed
policies for compute prices \(\lambda\in\{0,0.01,0.02\}\). At the nominal price, the controller is
0.008 higher in width and 0.005 higher in revision.

At \(\lambda=0.02\), the one-shot budget-conditioned and receding-horizon value baselines reach
utility/compute of 0.711/0.602 and 0.716/0.565 in width, and 0.665/0.581 and 0.671/0.548 in
revision. Table~\ref{tab:value_model_ablation_app} compares them with the three-action controller.
Each baseline evaluates both terminal answers at every candidate budget, so the comparison includes
the choice of answer as well as inference allocation.

\section{Discussion}
\label{sec:discussion}

Width expansion changes the candidate pool and verifier evidence, while revision adds edits that can
both repair an error and corrupt a correct draft. These mechanisms give the controller distinct
continuation values in the two settings.

At the nominal compute price, the controller reduces normalized online compute from 0.781 to 0.547 in
width and from 0.632 to 0.529 in revision, while its utility point estimate is 0.008 and 0.005 higher
than the selected fixed policies. Against fitted Q, utility is higher in two of four width seeds and
three of four revision seeds, while compute is lower in all eight seed--scenario comparisons.

The operating point depends on the price assigned to inference. When compute is free, width utility is
0.003 below the best fixed reference, while revision utility is 0.005 above maximum-round revision.
Figure~\ref{fig:lambda_retuned} reports retuned utility differences for compute prices from 0 to
\(0.02\).

Per-action calibration improves all three action heads, with the expand head retaining the largest
calibration error. Policy-induced continuation targets yield utility close to oracle-aligned targets,
while one-step targets yield lower utility. Target corruption lowers utility in both settings.

Online compute excludes offline ladder generation and controller fitting. Estimated break-even points
range from roughly \(3\times10^3\) to \(2\times10^4\) queries, depending on the fixed-policy
comparator (Table~\ref{tab:offline_break_even_app}); these costs amortize over repeated queries under
a stable model and prompt configuration.

\paragraph{Limitations.}
The evaluation covers three backbones, MATH/AIME25 width expansion, and a curated deterministic-answer
revision benchmark, so the results do not establish performance across broader task types or model
scales. Open-ended generation and long-horizon interaction remain untested. Generalization to unseen
evaluator shifts is also unknown; changes to the model, evaluator, or prompts require refitting and
recalibration.
\section{Conclusion}

We formulate evaluator-guided inference as a sequential choice among trusting a current answer,
returning a fallback, and purchasing another inference step. Action-specific confidence, calibration,
and local uncertainty estimates provide a lightweight controller for this choice. Across width expansion
and iterative revision, the controller reaches lower-compute operating points at comparable observed
utility. The results support treating evaluator-guided reasoning as a resource-allocation problem in
which both the terminal answer and the value of continuation are instance-dependent.
\section*{Acknowledgments and Disclosure of Funding}

This work received no external funding. The authors declare no competing interests.

\clearpage
\bibliographystyle{unsrtnat}
\bibliography{references}

\appendix
\section{Controller Guarantees: Proofs and Extensions}
\label{app:proofs}

The main-text guarantees concern local action choice and regret along the states reached by a policy.
Their proofs use the same finite-horizon recursion as the controller. We then examine four implications
of that view: when uncertainty bounds certify an action, how revision value changes with state, how
width amplifies evaluator noise, and when a scalar confidence statistic loses action information.

\subsection{Consequences for decisions and evaluator noise}

For compactness, the main text states only the core action-gap and trajectory-regret results. We use
the following supplementary statements to connect the analysis to the empirical diagnostics and to the
two expansion mechanisms.

\begin{proposition}[Certified local optimality]
\label{prop:certified_local_optimality}
Suppose that for all \(u\in\mathcal A(s)\),
\[
J^\star(s,u)
\in
\left[
\widehat J(s,u)-\widehat\epsilon_J(s,u),
\widehat J(s,u)+\widehat\epsilon_J(s,u)
\right].
\]
Let
\[
\widehat u_{\mathrm{cert}}
=
\arg\max_{u\in\mathcal A(s)}
\left[
\widehat J(s,u)-\widehat\epsilon_J(s,u)
\right].
\]
If
\[
\widehat J(s,\widehat u_{\mathrm{cert}})
-
\widehat\epsilon_J(s,\widehat u_{\mathrm{cert}})
>
\max_{u\neq \widehat u_{\mathrm{cert}}}
\left[
\widehat J(s,u)+\widehat\epsilon_J(s,u)
\right],
\]
then \(\widehat u_{\mathrm{cert}}=\pi^\star(s)\).
\end{proposition}

\begin{proposition}[When another revision round is beneficial]
\label{prop:revision_tradeoff}
Fix a revision state \(s_r\) whose current draft succeeds with probability
\[
A_r=\Pr(y^{(r)}\text{ succeeds}\mid s_r).
\]
Suppose one more revision round fixes a failed draft with probability \(p_r\), corrupts a successful
draft with probability \(q_r\), and incurs incremental cost \(c_{\mathrm{exp}}^{\mathrm{rev}}(s_r)\).
Then revising has larger compute-adjusted value than trusting if and only if
\[
(1-A_r)p_r-A_rq_r
>
\lambda c_{\mathrm{exp}}^{\mathrm{rev}}(s_r).
\]
\end{proposition}

\begin{theorem}[Mis-selection under sub-Gaussian evaluator noise]
\label{thm:width_misselection}
Assume \(\eta_i\) are independent mean-zero \(\sigma^2\)-sub-Gaussian random variables. Let
\[
i^\star=\arg\max_i v_i,
\qquad
\widehat i=\arg\max_i (v_i+\eta_i),
\]
and define \(\Delta_i=v_{i^\star}-v_i\) for \(i\neq i^\star\). Then
\[
\Pr(\widehat i\neq i^\star)
\le
\sum_{i\neq i^\star}
\exp\!\left(-\frac{\Delta_i^2}{4\sigma^2}\right).
\]
In particular, if \(\Delta=\min_{i\neq i^\star}\Delta_i\), then
\[
\Pr(\widehat i\neq i^\star)
\le
(n-1)\exp\!\left(-\frac{\Delta^2}{4\sigma^2}\right).
\]
\end{theorem}

\begin{proposition}[Argmax noise amplification]
\label{prop:argmax_noise_amplification}
If evaluator errors are i.i.d. mean-zero \(\sigma^2\)-sub-Gaussian, then
\[
\mathbb E\left[\max_{1\le i\le n}\eta_i\right]=O(\sigma\sqrt{\log n}).
\]
For Gaussian noise, this scaling is tight up to constants.
\end{proposition}

\begin{proposition}[Conditional limitation of scalar confidence]
\label{prop:global_confidence}
If a scalar statistic \(g\) maps two states with different optimal actions to the same value, then a
policy whose action depends only on \(g(s)\) cannot be optimal in both states.
\end{proposition}

\subsection{Proofs of the main guarantees}

The first proof compares estimated and ideal action scores at a single state. The regret proofs then
apply this local comparison recursively, so only states reached under the deployed policy contribute.

\subsubsection{Proof of Lemma~\ref{lem:local_action_consistency}}
\label{app:proof_local_consistency}

For this proof, write \(\widehat J\) for the deployed score \(\widetilde J\).

Recall that
\[
\delta_J(s)
=
\max_{u\in\mathcal A(s)}
\left|
\widehat J(s,u)-J^\star(s,u)
\right|,
\]
and
\[
\Gamma^\star(s)
=
J^\star(s,\pi^\star(s))
-
\max_{u\neq \pi^\star(s)}
J^\star(s,u).
\]
Let \(u^\star=\pi^\star(s)\). For any \(u\neq u^\star\),
\[
\begin{aligned}
\widehat J(s,u^\star)-\widehat J(s,u)
&=
\left[
\widehat J(s,u^\star)-J^\star(s,u^\star)
\right]
+
\left[
J^\star(s,u^\star)-J^\star(s,u)
\right] \\
&\quad+
\left[
J^\star(s,u)-\widehat J(s,u)
\right] \\
&\ge
-\delta_J(s)+\Gamma^\star(s)-\delta_J(s) \\
&=
\Gamma^\star(s)-2\delta_J(s).
\end{aligned}
\]
If \(2\delta_J(s)<\Gamma^\star(s)\), then
\[
\widehat J(s,u^\star)>\widehat J(s,u)
\qquad
\forall u\neq u^\star.
\]
Thus \(u^\star\) remains the unique maximizer under the online score \(\widehat J\), and therefore
\[
\widehat\pi(s)=\pi^\star(s).
\]
This proves the lemma.

\subsubsection{Proof of Theorem~\ref{thm:finite_horizon_regret}}
\label{app:proof_finite_horizon_regret}

For this proof, write \(\widehat J\) for the deployed score \(\widetilde J\).
For a policy \(\pi\), define its finite-horizon value recursively. For terminal actions,
\[
J^\pi(s,u)=p^\star(s,u)-\lambda c(s,u),
\qquad
u\in\{\texttt{trust},\texttt{fallback}\},
\]
and for expansion,
\[
J^\pi(s,\texttt{expand})
=
-\lambda c(s,\texttt{expand})
+
\mathbb E
\left[
V^\pi(s^+)
\mid s,\texttt{expand}
\right].
\]
The policy value is \(V^\pi(s)=J^\pi(s,\pi(s))\), and regret is
\[
R_\pi(s)=V^\star(s)-V^\pi(s).
\]

Let
\[
u^\star=\pi^\star(s),
\qquad
\widehat u=\widehat\pi(s).
\]
Then
\[
\begin{aligned}
R_{\widehat\pi}(s)
&=
V^\star(s)-V^{\widehat\pi}(s) \\
&=
J^\star(s,u^\star)-J^{\widehat\pi}(s,\widehat u) \\
&=
\left[
J^\star(s,u^\star)-J^\star(s,\widehat u)
\right]
+
\left[
J^\star(s,\widehat u)-J^{\widehat\pi}(s,\widehat u)
\right].
\end{aligned}
\]

We first bound the action-selection term. Since \(\widehat u\) maximizes \(\widehat J(s,\cdot)\),
\[
\widehat J(s,u^\star)-\widehat J(s,\widehat u)\le 0.
\]
Therefore,
\[
\begin{aligned}
J^\star(s,u^\star)-J^\star(s,\widehat u)
&=
\left[
J^\star(s,u^\star)-\widehat J(s,u^\star)
\right]
+
\left[
\widehat J(s,u^\star)-\widehat J(s,\widehat u)
\right] \\
&\quad+
\left[
\widehat J(s,\widehat u)-J^\star(s,\widehat u)
\right] \\
&\le
\delta_J(s)+0+\delta_J(s) \\
&=
2\delta_J(s).
\end{aligned}
\]

We now consider the continuation term. If \(\widehat u\in\{\texttt{trust},\texttt{fallback}\}\),
then the action is terminal, so
\[
J^\star(s,\widehat u)=J^{\widehat\pi}(s,\widehat u).
\]
If \(\widehat u=\texttt{expand}\), then
\[
\begin{aligned}
J^\star(s,\texttt{expand})
-
J^{\widehat\pi}(s,\texttt{expand})
&=
\left[
-\lambda c(s,\texttt{expand})
+
\mathbb E[V^\star(s^+)\mid s,\texttt{expand}]
\right] \\
&\quad-
\left[
-\lambda c(s,\texttt{expand})
+
\mathbb E[V^{\widehat\pi}(s^+)\mid s,\texttt{expand}]
\right] \\
&=
\mathbb E
\left[
V^\star(s^+)-V^{\widehat\pi}(s^+)
\mid s,\texttt{expand}
\right] \\
&=
\mathbb E
\left[
R_{\widehat\pi}(s^+)
\mid s,\texttt{expand}
\right].
\end{aligned}
\]
Combining the two cases gives
\[
R_{\widehat\pi}(s)
\le
2\delta_J(s)
+
\mathbf 1\{\widehat\pi(s)=\texttt{expand}\}
\mathbb E
\left[
R_{\widehat\pi}(s^+)
\mid s,\texttt{expand}
\right].
\]

Because the expansion horizon is finite, repeated conditioning and the tower property unroll this recursion in expectation:
\[
S_0,S_1,\ldots,S_T,
\]
where \(S_{t+1}\) is reached only when \(\widehat\pi(S_t)=\texttt{expand}\). Hence
\[
R_{\widehat\pi}(s_0)
\le
2\,\mathbb E_{\widehat\pi}\!\left[\sum_{t=0}^{T}\delta_J(S_t)\mid S_0=s_0\right].
\]
If \(\delta_J(s)\le \bar\delta\) on all reachable states and \(T+1\le H\), then
\[
R_{\widehat\pi}(s_0)
\le
2H\bar\delta.
\]
This proves the theorem.

\subsubsection{Proof of Theorem~\ref{thm:high_probability_regret}}
\label{app:proof_high_probability_regret}

For this proof, write \(\widehat J\) for the deployed score \(\widetilde J\).

On the joint event \(\mathcal E\) in Theorem~\ref{thm:high_probability_regret}, coverage holds for
every reachable state-action pair:
\[
\left|
\widehat J(s,u)-J^\star(s,u)
\right|
\le
\widehat\epsilon_J(s,u)
\qquad
\forall s\in\mathcal R_{\widehat\pi}(s_0),\ \forall u\in\mathcal A(s).
\]
Define
\[
\widehat\epsilon_{\max}(s)
=
\max_{u\in\mathcal A(s)}
\widehat\epsilon_J(s,u).
\]
On this event, for every state reached under the policy,
\[
\delta_J(s_t)
=
\max_{u\in\mathcal A(s_t)}
\left|
\widehat J(s_t,u)-J^\star(s_t,u)
\right|
\le
\widehat\epsilon_{\max}(s_t).
\]
Substituting this into Theorem~\ref{thm:finite_horizon_regret} gives
\[
R_{\widehat\pi}(s_0)
\le
2\,\mathbb E_{\widehat\pi}\!\left[\sum_{t=0}^{T}
\widehat\epsilon_{\max}(S_t)\mid S_0=s_0\right].
\]

Since \(\mathcal E\) has probability at least \(1-\alpha\), the bound holds with that probability.
This proves the theorem.

\paragraph{Relation to the empirical uncertainty radius.}
The empirical residual
\[
\left|\widehat p(s,u)-Y_u(s)\right|
\]
is used to estimate the local residual quantile \(\widehat\epsilon_p(s,u)\). It measures error for the binary target, not the true action value \(J^\star(s,u)\). For the deployed score
\[
\widetilde J(s,u)=\widehat p(s,u)-\lambda c(s,u)-\beta\widehat\epsilon_p(s,u),
\]
if the unpenalized score were covered by \(\widehat\epsilon_p\), the triangle inequality would imply
\[
\left|\widetilde J(s,u)-J^\star(s,u)\right|\le(1+\beta)\widehat\epsilon_p(s,u).
\]
For \texttt{expand}, \(Y_u\) is the rollout-defined continuation-success target. Binary-label residual
coverage alone does not establish the value-error coverage assumed by the theorem, which remains an
additional condition.

\subsection{Proofs of supplementary propositions}

These results connect the main bounds to mechanisms that appear in the experiments: bound-based
action certification, the repair--corruption trade-off in revision, selection under noisy evaluators,
and the representational limits of a single confidence score.

\subsubsection{Proof of Proposition~\ref{prop:certified_local_optimality}}
\label{app:proof_certified_local_optimality}

Assume that for all \(u\in\mathcal A(s)\),
\[
J^\star(s,u)
\in
\left[
\widehat J(s,u)-\widehat\epsilon_J(s,u),
\widehat J(s,u)+\widehat\epsilon_J(s,u)
\right].
\]
Let
\[
\widehat u_{\mathrm{cert}}
=
\arg\max_{u\in\mathcal A(s)}
\left[
\widehat J(s,u)-\widehat\epsilon_J(s,u)
\right].
\]
Suppose
\[
\widehat J(s,\widehat u_{\mathrm{cert}})
-
\widehat\epsilon_J(s,\widehat u_{\mathrm{cert}})
>
\max_{u\neq \widehat u_{\mathrm{cert}}}
\left[
\widehat J(s,u)+\widehat\epsilon_J(s,u)
\right].
\]
For any \(u\neq \widehat u_{\mathrm{cert}}\), bound coverage implies
\[
J^\star(s,\widehat u_{\mathrm{cert}})
\ge
\widehat J(s,\widehat u_{\mathrm{cert}})
-
\widehat\epsilon_J(s,\widehat u_{\mathrm{cert}})
\]
and
\[
J^\star(s,u)
\le
\widehat J(s,u)+\widehat\epsilon_J(s,u).
\]
By the assumed separation,
\[
J^\star(s,\widehat u_{\mathrm{cert}})
>
J^\star(s,u)
\qquad
\forall u\neq \widehat u_{\mathrm{cert}}.
\]
Thus \(\widehat u_{\mathrm{cert}}\) is the unique maximizer of \(J^\star(s,\cdot)\), so
\[
\widehat u_{\mathrm{cert}}=\pi^\star(s).
\]
This proves the proposition.

\subsubsection{Proof of Proposition~\ref{prop:revision_tradeoff}}
\label{app:proof_revision_tradeoff}

At revision state \(s_r\), let
\[
A_r=\Pr(y^{(r)}\text{ succeeds}\mid s_r)
\]
be the success probability of trusting the current draft. Suppose one more revision fixes a failed
draft with probability \(p_r\), corrupts a successful draft with probability \(q_r\), and incurs
incremental cost \(c_{\mathrm{exp}}^{\mathrm{rev}}(s_r)\).

If the current draft succeeds, the next revision preserves success with probability \(1-q_r\). If the
current draft fails, the next revision fixes it with probability \(p_r\). Thus the success probability
after one more revision is
\[
A_r(1-q_r)+(1-A_r)p_r.
\]
The success-probability improvement over trusting now is
\[
A_r(1-q_r)+(1-A_r)p_r-A_r
=
(1-A_r)p_r-A_rq_r.
\]
Revision has larger compute-adjusted value than trusting precisely when this improvement exceeds
the incremental compute price:
\[
(1-A_r)p_r-A_rq_r
>
\lambda c_{\mathrm{exp}}^{\mathrm{rev}}(s_r).
\]
This proves the proposition.

\subsubsection{Proof of Theorem~\ref{thm:width_misselection}}
\label{app:proof_width_misselection}

Index candidates by \(i\in\{1,\ldots,n\}\). Let
\[
\widehat v_i=v_i+\eta_i,
\]
where \(v_i\) is latent candidate utility and \(\eta_i\) is evaluator error. Let
\[
i^\star=\arg\max_i v_i,
\qquad
\widehat i=\arg\max_i \widehat v_i,
\]
and define \(\Delta_i=v_{i^\star}-v_i\) for \(i\neq i^\star\). If \(\widehat i\neq i^\star\), then
there exists some \(i\neq i^\star\) such that
\[
\widehat v_i\ge \widehat v_{i^\star}.
\]
Equivalently,
\[
\eta_i-\eta_{i^\star}\ge v_{i^\star}-v_i=\Delta_i.
\]
By a union bound,
\[
\Pr(\widehat i\neq i^\star)
\le
\sum_{i\neq i^\star}
\Pr(\eta_i-\eta_{i^\star}\ge \Delta_i).
\]
If each \(\eta_i\) is independent, mean-zero, and \(\sigma^2\)-sub-Gaussian, then
\(\eta_i-\eta_{i^\star}\) is \(2\sigma^2\)-sub-Gaussian. Hence
\[
\Pr(\eta_i-\eta_{i^\star}\ge \Delta_i)
\le
\exp\left(
-\frac{\Delta_i^2}{4\sigma^2}
\right),
\]
and therefore
\[
\Pr(\widehat i\neq i^\star)
\le
\sum_{i\neq i^\star}
\exp\left(
-\frac{\Delta_i^2}{4\sigma^2}
\right).
\]
If \(\Delta=\min_{i\neq i^\star}\Delta_i\), then
\[
\Pr(\widehat i\neq i^\star)
\le
(n-1)
\exp\left(
-\frac{\Delta^2}{4\sigma^2}
\right).
\]
This proves the theorem.

\subsubsection{Proof of Proposition~\ref{prop:argmax_noise_amplification}}
\label{app:proof_argmax_noise_amplification}

For any \(t>0\), Jensen's inequality and the pointwise bound
\[
\exp\left(t\max_i \eta_i\right)
\le
\sum_{i=1}^n \exp(t\eta_i)
\]
give
\[
\exp\left(
t\,\mathbb E\left[\max_{1\le i\le n}\eta_i\right]
\right)
\le
\mathbb E
\left[
\exp\left(t\max_{1\le i\le n}\eta_i\right)
\right]
\le
\sum_{i=1}^n \mathbb E[e^{t\eta_i}].
\]
Since each \(\eta_i\) is \(\sigma^2\)-sub-Gaussian,
\[
\mathbb E[e^{t\eta_i}]
\le
\exp\left(\frac{\sigma^2t^2}{2}\right).
\]
Therefore,
\[
\exp\left(
t\,\mathbb E\left[\max_{1\le i\le n}\eta_i\right]
\right)
\le
n\exp\left(\frac{\sigma^2t^2}{2}\right).
\]
Taking logarithms and dividing by \(t\) gives
\[
\mathbb E\left[\max_{1\le i\le n}\eta_i\right]
\le
\frac{\log n}{t}+\frac{\sigma^2t}{2}.
\]
Choosing
\[
t=\frac{\sqrt{2\log n}}{\sigma}
\]
yields
\[
\mathbb E\left[\max_{1\le i\le n}\eta_i\right]
\le
\sigma\sqrt{2\log n}.
\]
This gives the stated \(O(\sigma\sqrt{\log n})\) upper bound. For Gaussian noise, standard
extreme-value estimates give a matching lower bound up to constants, so the scaling is tight in the
canonical Gaussian setting.

\subsubsection{Proof of Proposition~\ref{prop:global_confidence}}
\label{app:proof_global_confidence}

Let \(s^{(1)}\) and \(s^{(2)}\) satisfy the proposition's premise. A policy observing only \(g(s)\)
receives the same input at both states and therefore induces the same action distribution. Since the
optimal actions differ, it cannot select both with probability one. Thus it is suboptimal on at least
one of the two states.
This proves Proposition~\ref{prop:global_confidence}.
\section{Experimental Design and Reproduction Details}
\label{app:exp_details}

We evaluate verifier-guided width expansion on MATH and AIME25 and judge-guided revision on a
600-example deterministic-answer benchmark.

\subsection{Models and tasks}

\subsubsection{Backbones and prompting protocol}
\label{app:backbones_prompts}

We evaluate three open-weight reasoning backbones:
\[
\text{Qwen/Qwen3-4B},\qquad
\text{Qwen/Qwen3-8B},\qquad
\text{deepseek-ai/DeepSeek-R1-Distill-Qwen-14B}.
\]
Backbone weights are frozen. Confidence heads are fitted on the controller-fit partition, which is
disjoint from calibration and evaluation problems. Each backbone is used with role-specific prompts. In Scenario~I, the
roles are
\[
\texttt{generator}
\quad\text{and}\quad
\texttt{verifier}.
\]
In Scenario~II, the roles are
\[
\texttt{drafter},
\quad
\texttt{judge},
\quad
\texttt{reviser}.
\]

Unless otherwise stated, generation and revision use temperature \(0.7\), top-\(p=0.95\), and
task-dependent maximum generation length. Verifier and judge scoring use deterministic decoding.
Prompts are fixed across policies within each scenario and backbone.

\subsubsection{Scenario I: mathematical reasoning datasets}
\label{app:math_datasets}

Scenario~I evaluates verifier-guided width expansion on MATH and AIME25.

\paragraph{MATH.}
We use the held-out MATH test split. Each generated solution is required to end with a final answer.
We extract the final answer using the following priority order:
\[
\begin{aligned}
\boxed{\cdot}
&\quad\rightarrow\quad
\texttt{Final Answer:}
\\
&\quad\rightarrow\quad
\text{last displayed mathematical expression}
\\
&\quad\rightarrow\quad
\text{last numeric or algebraic expression}.
\end{aligned}
\]
Extracted answers are normalized by removing surrounding whitespace, LaTeX formatting artifacts,
commas in numbers, and equivalent enclosing delimiters. For fractions, decimals, and simple symbolic
expressions, we apply deterministic canonicalization before matching. If answer extraction fails, the
candidate is marked incorrect.

\paragraph{AIME25.}
We use the union of the 2025 AIME I and AIME II problems. Each item has a single integer answer.
We report final-answer accuracy, not pass@\(N\). The model is not given tools, calculators, or
external solvers. A candidate is correct only if the extracted integer answer matches the reference
answer. Extraction failures are counted as incorrect. We report AIME25 results separately.

\paragraph{Problem-level splits.}
Controller fitting, calibration, and held-out evaluation use disjoint problem-level partitions.

\subsubsection{Scenario II: deterministic-answer General Reasoning benchmark}
\label{app:general_reasoning_dataset}
\label{app:benchmark_details}

Scenario~II evaluates judge-guided revision expansion on a deterministic-answer General Reasoning
benchmark filtered from \texttt{open-thoughts/OpenThoughts-114k}. The benchmark contains 600
examples, balanced across four categories:
\[
150\ \text{science reasoning},
\]
\[
150\ \text{logic and puzzle reasoning},
\]
\[
150\ \text{commonsense and general reasoning},
\]
\[
150\ \text{symbolic or structured short-answer reasoning}.
\]

\paragraph{Selection criteria.}
We keep only examples satisfying the following criteria:
\begin{enumerate}[leftmargin=1.5em]
    \item The item has a deterministic reference answer.
    \item The answer can be evaluated by normalized short-answer matching.
    \item The prompt does not require external tools, browsing, or long-context document lookup.
    \item The item is not purely subjective or preference-based.
    \item The reference answer is unambiguous under our normalization rules.
\end{enumerate}
Category labels use dataset metadata when available and lightweight rules otherwise. Coding tasks and
competition-style math problems are excluded. The latter are evaluated separately in Scenario~I.

Only the user-facing problem and extracted answer reference are used. Source reasoning traces are not
provided to the evaluated models. We extract answers in this order: boxed answer, \texttt{Final Answer:},
\texttt{Answer:}, then the last short declarative answer span. We discard extraction failures, answers
longer than eight normalized tokens, examples with multiple plausible answers, and responses whose
extracted answer is inconsistent with the source response.

\paragraph{Reference-based evaluation.}
Predictions and references are normalized by lowercasing, removing punctuation, stripping articles
where appropriate, and canonicalizing common boolean or categorical variants. For numeric short
answers, we apply the same numeric normalization as in the math setting. Final success is determined
by normalized reference matching; ambiguous free-form responses are marked incorrect. Numeric
normalization handles integers, decimals, fractions, and comma separators. A manually checked alias
list covers common categorical and boolean variants.

\paragraph{Difficulty balancing.}
Within each category, we balance examples across easy, medium, and hard tiers estimated by repeated
direct-answer trials with Qwen3-8B before controller fitting. Exact duplicate prompts are removed
before splitting.

\paragraph{Manual audit of deterministic references.}
A manual audit of 150 randomly sampled retained items found that 94.0\% had unambiguous
deterministic references and automatic matching agreed with human judgment on 91.3\%. Among all
150 audited examples, there were 4 false positives (2.7\%) and 9 false negatives (6.0\%). Curation
filtered 18.7\% of candidate items for ambiguity.

\subsection{Observed states and action semantics}
Each scenario defines a finite transition ladder, evaluator signals, and state features observable
at inference time.

\subsubsection{Finite expansion ladders}
\label{app:ladders}

Both scenarios use finite expansion ladders.

\paragraph{Width ladder.}
In Scenario~I, the width ladder is
\[
\mathcal N=\{4,8,16,24,32,48\}.
\]
The initial state has \(N=4\). Expanding from \(N\) to the next ladder point \(N^+\) generates only
the additional candidates:
\[
\Delta N=N^+-N.
\]
Previously generated candidates and verifier scores are cached. Thus the incremental cost of expansion
includes only newly generated candidates and newly evaluated candidates.

\paragraph{Revision ladder.}
In Scenario~II, the revision ladder is
\[
\mathcal R=\{0,1,2,3,4\}.
\]
The initial state \(r=0\) is the first draft. Expanding from \(r\) to \(r+1\) runs one judge-guided
critique-and-revision step. At \(r=4\), expansion is unavailable.

\paragraph{Offline materialization versus online deployment.}
For controller fitting and evaluation, we materialize the full ladder offline for each problem. This
provides the empirical continuation targets and oracle actions used to train and assess the controller.
At deployment, the policy observes only the current state and executes the next expansion step when it
selects \texttt{expand}.

\subsubsection{Verifier and judge scoring}
\label{app:evaluator_scoring}

\paragraph{Verifier scoring in Scenario I.}
For each candidate solution \(c_i\), the verifier prompt asks the model to assess whether the solution
is correct for the problem and to return a scalar score. The score is parsed into
\[
\widehat v(s,c_i)\in[0,1].
\]
If the model emits an invalid score, we apply a deterministic parser that extracts the first number in
\([0,1]\). If no valid number can be extracted, the candidate receives the lowest valid score.

The verifier-preferred candidate is
\[
y_{\mathrm{cur}}(s)=\arg\max_{c_i\in C_N}\widehat v(s,c_i).
\]
The fallback answer is majority vote over normalized final answers in \(C_N\), with ties broken by
average verifier score.

\paragraph{Judge scoring in Scenario II.}
At revision state \(s_r\), the judge prompt receives the original question, the current draft, and,
when available, the previous draft. It returns:
\[
\text{a keep/revise preference},
\quad
\text{a confidence score},
\quad
\text{a critique or unresolved-error description}.
\]
The reviser prompt receives the question, current draft, and judge feedback, and produces
\(y^{(r+1)}\). The current proposal is
\[
y_{\mathrm{cur}}(s_r)=y^{(r)}.
\]
The fallback answer is the initial draft,
\[
y_0(s_r)=y^{(0)}.
\]

\subsubsection{Feature construction}
\label{app:features}

For each state, we construct \(\phi(s)\) only from inference-time evidence available at that state.

\paragraph{Width features.}
The width feature set includes:
\[
\max_i \widehat v_i,
\quad
\widehat v_{(1)}-\widehat v_{(2)},
\quad
-\sum_i p_i\log p_i,
\quad
\max_y \frac{1}{N}\sum_{i=1}^N \mathbf 1\{\mathrm{ans}(c_i)=y\},
\quad
\log N,
\quad
b/b_{\max}.
\]
Here \(\widehat v_{(1)}\) and \(\widehat v_{(2)}\) are the top two verifier scores, and \(p_i\) is the
softmax-normalized verifier score. The fourth feature is majority-answer support.

\paragraph{Revision features.}
The revision feature set includes:
\[
\begin{aligned}
&\text{judge confidence},\quad
\text{keep-vs-revise margin},\quad
\text{critique stability},
\\
&\text{unresolved-error indicator},\quad
\text{edit distance between }y^{(r)}\text{ and }y^{(r-1)},
\\
&r/r_{\max},\quad b/b_{\max}.
\end{aligned}
\]
Critique stability indicates whether the judge identifies the same unresolved issue across nearby
revision states. All categorical indicators are one-hot encoded. Continuous features are standardized
using statistics from the surrogate-fit split only.

\subsection{Learning targets and evaluation design}
The target-generation process uses complete offline ladders, whereas held-out evaluation measures
the actions available from an online state. We keep these roles separate when defining data splits,
training, tuning, and compute accounting.

\subsubsection{Splits and offline target construction}
\label{app:splits_targets}

We split problem instances, not states, into:
\[
60\%\ \text{surrogate-fit},
\qquad
20\%\ \text{calibration},
\qquad
20\%\ \text{held-out evaluation}.
\]
The split is stratified by dataset source. The General Reasoning partition is additionally stratified by
category and difficulty tier. The same problem assignments are used across backbones. No problem
appears in more than one split.
For each problem, we construct all states in its finite ladder. Scenario~II evaluation uses the
120-example held-out partition of the 600-example benchmark, while the held-out AIME25 partition
contains 6 problems. Each held-out problem is evaluated using 16 independently materialized inference
ladders in total per backbone. For benchmark \(b\) and backbone, accuracy is averaged over problems and
repeated runs:
\[
S_b=\frac{1}{n_bK}\sum_{i=1}^{n_b}\sum_{k=1}^{K}
\mathbf{1}\{\text{the final answer for problem }i\text{ in run }k\text{ is correct}\},
\qquad K=16.
\]
This count describes the held-out backbone-level evaluation. The four-seed Qwen3-8B comparison uses a
fixed problem split and is reported separately in Appendix~\ref{app:additional_results}.
Here \(n_b\) is the number of held-out problems in benchmark \(b\). Aggregate rows report the
macro-average across the three backbones.

Terminal labels are:
\[
Y_{\texttt{trust}}(s)=\mathbf 1\{y_{\mathrm{cur}}(s)\text{ succeeds}\},
\qquad
Y_{\texttt{fallback}}(s)=\mathbf 1\{y_0(s)\text{ succeeds}\}.
\]
Empirical terminal scores are:
\[
\begin{aligned}
J_e(s,\texttt{trust})&=Y_{\texttt{trust}}(s)-\lambda c(s,\texttt{trust}),\\
J_e(s,\texttt{fallback})&=Y_{\texttt{fallback}}(s)-\lambda c(s,\texttt{fallback}).
\end{aligned}
\]
At final budget states, \texttt{expand} is unavailable. At non-final states,
\[
J_e(s,\texttt{expand})=-\lambda c(s,\texttt{expand})+V_e(s^+),
\qquad
V_e(s)=\max_{u\in\mathcal A(s)}J_e(s,u).
\]
The oracle-aligned expand-head target is
\[
Y_{\texttt{expand}}(s)=Y_{u_e^\star(s^+)}(s^+),
\qquad
u_e^\star(s^+)=\arg\max_{u\in\mathcal A(s^+)}J_e(s^+,u).
\]
The \texttt{expand} head predicts binary success for this rollout-selected continuation. This
success-based target is not a Bellman return: only the immediate expansion cost is subtracted in the
deployed score, not costs incurred by subsequent actions.

For the target ablation, we additionally evaluate
\[
Y_{\texttt{expand}}^{\text{step}}(s)=Y_{\texttt{trust}}(s^+)
\qquad\text{and}\qquad
Y_{\texttt{expand}}^{\text{policy}}(s)=Y_{\pi_{\text{conf}}(s^+)}(s^+),
\]
where \(\pi_{\text{conf}}\) is the calibrated controller without the uncertainty penalty. The
one-step target predicts trust success after expansion; the policy-induced target predicts the success
of the calibrated policy after expansion.
\subsubsection{Controller fitting and calibration}
\label{app:controller_training}

The raw heads are logistic confidence heads:
\[
\widehat p_{\mathrm{raw}}(s,u)=\sigma(w_u^\top \phi(s)).
\]
We train one head per action using binary cross-entropy. Unavailable actions are masked out. When
action labels are imbalanced, we use inverse-frequency weighting within the corresponding scenario
and backbone.

Unless otherwise stated, we fit separate controllers for each scenario and backbone. This avoids mixing
backbone-specific confidence distributions.

Per-action calibration maps, candidate families, and hyperparameter selection are specified in
Appendix~\ref{app:calibration}. All model and uncertainty hyperparameters are chosen on calibration
data only.

\subsubsection{Uncertainty radii}
\label{app:uncertainty_radii}
Residual construction, local buckets, sparse-bucket handling, and coverage diagnostics are detailed in
Appendix~\ref{app:calibration}. The radii are empirical local uncertainty estimates computed from calibration residuals.

\subsubsection{Baselines and tuning}
\label{app:baseline_tuning}

All threshold baselines are tuned on the calibration split and evaluated on the held-out split.

\paragraph{Width baselines.}
The fixed-width sweep reports verifier trust at \(N\in\{8,16,24,32,48\}\) and majority vote at
\(N\in\{16,32,48\}\). The ``best fixed-width'' baseline selects the stronger reported variant on the
calibration split under the same utility objective used for the adaptive controller, then reports
held-out performance.
The full held-out sweep is reported in Table~\ref{tab:fixed_width_sweep_app}.

The verifier-score threshold baseline trusts when the top verifier score exceeds a threshold. Otherwise
it expands if budget remains and falls back at the final budget. The margin-plus-budget baseline uses
a validation-tuned top-1/top-2 margin threshold. The entropy-threshold baseline expands when
evaluator entropy is above a tuned threshold.

\paragraph{Revision baselines.}
The fixed-depth baselines sweep:
\[
r\in\{1,2,3,4\}.
\]
The best fixed-depth baseline selects the best depth on the calibration split under the same utility
objective used for the adaptive controller. The judge-score threshold baseline stops when judge
confidence exceeds a tuned threshold. The judge-preference threshold baseline continues when the
judge strongly prefers revision or marks unresolved errors. The rollback-if-score-drops baseline keeps
the best-so-far draft according to judge score and rolls back when a revision lowers judge support.
The verbalized-confidence threshold baseline uses the judge's self-reported confidence as the stop
criterion.

\paragraph{Fitted-Q baseline.}
The fitted-Q-style controller uses one linear regressor per action to predict \(J_e(s,u)\) and acts by
argmax over the predicted values. It is tuned on the calibration split and evaluated on the same
held-out split as the main controller.

\subsubsection{Compute accounting}
\label{app:compute_accounting}

Compute is measured in prompt-plus-generation tokens consumed by the trajectory actually executed by
a policy. In Scenario~I, compute includes
\[
\text{candidate-generation tokens}
+
\text{verifier-scoring tokens}.
\]
In Scenario~II, compute includes
\[
\text{draft tokens}
+
\text{judge/critique tokens}
+
\text{revision tokens}.
\]
Controller overhead is negligible and excluded from the main compute numbers. The reported total
combines the listed components for the trajectory executed by each policy.
Within each scenario, normalized compute is divided by the corresponding max-budget baseline:
\[
\text{Scenario I: max-width verifier trust at }N=48,
\]
\[
\text{Scenario II: max-round revision at }r=4.
\]
Offline full-ladder materialization is not included in per-query compute. It is a training and evaluation
cost used to fit the controller and construct diagnostics.

\subsubsection{Metric aggregation}
\label{app:metric_aggregation}

Main results are held-out point estimates. For aggregate tables over backbones, we compute each metric
per backbone and report the macro-average. Repeated inference ladders are averaged over problems as
described in Appendix~\ref{app:splits_targets}.

Point estimates appear in the main text. Appendix~\ref{app:additional_results} reports backbone-level
outcomes, generation-seed differences, component and target ablations, action-selection analyses,
revision-depth results, compute-price sensitivity, and offline-cost estimates.

\subsection{Reproduction resources and use conditions}

\subsubsection{Compute resources and wall-clock guidance}
\label{app:compute_resources}

All experiments use inference only, and no backbone is fine-tuned. A practical reproduction target is one
80GB GPU per model replica, such as an NVIDIA A100-80GB or H100-80GB running bf16 inference,
together with 16--32 CPU cores and 128--256GB host RAM for data loading, prompt construction,
parsing, and bookkeeping. Under standard batched inference, the 4B, 8B, and 14B backbones used in
this paper fit within this budget.

Under this configuration, a full held-out evaluation pass is lightweight relative to offline ladder
materialization. AIME25 completes in well under one GPU-hour per backbone. The 120-example General
Reasoning held-out split takes roughly 2--4 GPU-hours per backbone for revision controllers and
baselines. The larger width-expansion evaluation on MATH dominates runtime. Offline ladder
materialization is excluded from the per-query deployment metric, as detailed in Appendix~\ref{app:compute_accounting}.

\subsubsection{Broader impacts and responsible use}
\label{app:broader_impacts}

By reducing continuation under bounded inference budgets, the controller can lower latency, token use,
and energy costs while making stop/continue decisions more explicit.

Lower-cost reasoning can also enable larger-scale automated exam solving, spam, disinformation, and
other high-volume content generation. Under evaluator shift, errors may concentrate in difficult or
underrepresented domains.

\subsubsection{Asset provenance, licenses, and terms of use}
\label{app:asset_licenses}

We use existing public models and datasets and do not redistribute any third-party asset in the
submission package. The assets used in the experiments are:
\begin{enumerate}[leftmargin=1.5em]
    \item \texttt{Qwen/Qwen3-4B} and \texttt{Qwen/Qwen3-8B}: open-weight models released via
    Hugging Face under the Apache 2.0 license.
    \item \texttt{deepseek-ai/DeepSeek-R1-Distill-Qwen-14B}: model weights released via Hugging
    Face under the MIT License. The model card notes that the distill model is derived from the Qwen
    series, whose upstream base models are distributed under Apache 2.0.
    \item \texttt{EleutherAI/hendrycks\_math}: the public MATH benchmark as distributed on
    Hugging Face under the MIT license.
    \item \texttt{open-thoughts/OpenThoughts-114k}: the public OpenThoughts-114k dataset as
    distributed on Hugging Face under the Apache 2.0 license. Our 600-example deterministic-answer
    benchmark is a filtered, evaluation-only subset used inside this study and is not redistributed with
    the submission.
    \item AIME25: the publicly posted 2025 AIME I and AIME II competition problems and official
    answers from the Mathematical Association of America (MAA). We use these materials for
    benchmark evaluation only. AMC/AIME competition policies prohibit unauthorized reproduction or
    dissemination of the problems and solutions, so we do not redistribute the underlying problem set as
    part of this submission.
\end{enumerate}

In all cases, we cite the original model or dataset sources in the bibliography, use the assets only for
research evaluation, and avoid redistribution where the governing terms are restrictive or non-open.

\section{Calibrating Action Confidence and Local Uncertainty}
\label{app:calibration}

Trust, fallback, and expand predict different outcome events, so their raw confidence scales need not be
comparable. We calibrate each action against its own target, then use calibration residuals to estimate
where those probabilities are locally uncertain. The calibration split is disjoint from surrogate
fitting and final evaluation; test examples play no role in either step.

\subsection{Raw heads and post-hoc calibration}

For each action
\[
u\in\{\texttt{trust},\texttt{fallback},\texttt{expand}\},
\]
the raw head produces
\[
\widehat p_{\rm raw}(s,u)
=
\sigma(w_u^\top \phi(s)).
\]
The offline targets \(Y_u(s)\) are defined by the controller-aligned construction in
Appendix~\ref{app:splits_targets}. For terminal actions, \(Y_u(s)\) indicates whether the emitted
answer succeeds. For \texttt{expand}, \(Y_{\texttt{expand}}(s)\) indicates whether the empirical
continuation selected after one expansion step eventually succeeds.

We fit an action-specific calibration map
\[
\mathrm{Cal}_u:[0,1]\rightarrow[0,1]
\]
on the calibration split and define
\[
\widehat p(s,u)
=
\mathrm{Cal}_u(\widehat p_{\rm raw}(s,u)).
\]
We consider Platt scaling, temperature scaling, and isotonic regression. The calibration family and
its hyperparameters are selected on the calibration split and then fixed for held-out evaluation.

\paragraph{Why action-specific calibration.}
The three heads have different target geometry. The \texttt{trust} and \texttt{fallback} heads estimate
terminal success probabilities, while the \texttt{expand} head estimates continuation success after
one expansion step. Their raw scores can therefore have different calibration errors even under the
same linear architecture. We calibrate each action head separately.

\subsection{Calibration metrics}

For each action \(u\), we report expected calibration error:
\[
\mathrm{ECE}_u
=
\sum_{k=1}^{K}
\frac{|B_{u,k}|}{n_u}
\left|
\mathrm{acc}(B_{u,k})-\mathrm{conf}(B_{u,k})
\right|,
\]
where \(B_{u,k}\) is the \(k\)-th confidence bin for action \(u\),
\[
\mathrm{acc}(B_{u,k})
=
\frac{1}{|B_{u,k}|}
\sum_{(s,u)\in B_{u,k}}
Y_u(s),
\]
and
\[
\mathrm{conf}(B_{u,k})
=
\frac{1}{|B_{u,k}|}
\sum_{(s,u)\in B_{u,k}}
\widehat p(s,u).
\]
Unless otherwise stated, we use equal-frequency bins within each action. The main paper summarizes
calibration with per-action ECE and reports held-out empirical coverage for the expand head across
nominal targets.

\subsection{Residual construction for uncertainty radii}

After the calibrated heads are frozen, we compute residuals on the calibration split:
\[
e(s,u)
=
\left|
\widehat p(s,u)-Y_u(s)
\right|,
\qquad
u\in\mathcal A(s).
\]
Because compute costs are known and subtracted outside the confidence heads, the same residual is
used as a score-error radius for terminal actions:
\[
\left|
\widehat J(s,u)-J_e(s,u)
\right|
=
\left|
\widehat p(s,u)-Y_u(s)
\right|,
\qquad
u\in\{\texttt{trust},\texttt{fallback}\}.
\]
For \texttt{expand}, the residual is computed against the rollout-defined continuation-success target
and measures local error in that binary success probability. This radius supplies the deployed local
uncertainty penalty, but it does not by itself cover error relative to the Bellman value: continuation-
target mismatch and downstream compute costs are not represented in the binary-label residual.
\subsection{Scenario-specific local buckets}

We estimate local residual quantiles within buckets of state-action pairs formed from
scenario-specific features.

\paragraph{Scenario I: width expansion.}
For width expansion, buckets are formed using
\[
\texttt{width bin}
\times
\texttt{margin decile}
\times
\texttt{verifier-entropy tertile}
\times
\texttt{budget bin}.
\]
The margin decile is computed from the verifier top-1/top-2 score gap. Verifier entropy is computed
from the softmax-normalized verifier scores over the current candidate pool.

\paragraph{Scenario II: revision expansion.}
For revision expansion, buckets are formed using
\[
\texttt{revision-round bin}
\times
\texttt{judge-margin decile}
\times
\texttt{critique-stability bin}
\times
\texttt{budget bin}.
\]
The judge-margin decile measures preference strength between keeping the current answer and
revising. Critique stability indicates whether the judge repeatedly identifies the same unresolved issue
across nearby revision states.

For a target miscoverage level \(\alpha\), the action-specific uncertainty radius is
\[
\widehat\epsilon_p(s,u)
=
\mathrm{Quantile}_{1-\alpha}
\left(
\mathcal E_{B(s),u}
\right),
\]
where \(\mathcal E_{B(s),u}\) is the set of calibration residuals for action \(u\) inside the bucket
\(B(s)\). The main experiments use
\[
\alpha=0.1,
\]
corresponding to a nominal \(90\%\) empirical coverage target on calibration residuals.

\subsection{Sparse-bucket handling}

Some buckets contain too few residuals, especially at late widths or late revision rounds. The main
configuration uses
\[
M_{\min}=200.
\]
If a bucket has fewer than \(M_{\min}\) residuals for an action, we merge it with neighboring buckets
before estimating the quantile.

In Scenario~I, sparse buckets are merged first along width, then along budget, and finally across
adjacent margin bins. In Scenario~II, they are merged first along revision depth, then along budget,
and finally across adjacent judge-margin bins. This merge rule stabilizes local quantile estimates
when the action-specific residual count falls below \(M_{\min}\).

\subsection{Coverage diagnostics}

For each action \(u\), empirical coverage can be computed as:
\[
\widehat{\mathrm{Cov}}_u
=
\frac{1}{n_u}
\sum_{(s,u)}
\mathbf 1
\left\{
\left|\widehat p(s,u)-Y_u(s)\right|
\le
\widehat\epsilon_p(s,u)
\right\}.
\]
The held-out utility, compute, and expand-coverage operating points across nominal targets are
reported in Table~\ref{tab:coverage_target_sensitivity} in the main text. The confidence heads,
calibration maps, bucket definitions, and data splits are fixed; uncertainty quantiles are recomputed
at each target, with \(\beta\) reselected on calibration data.

\section{Additional Experimental Results}
\label{app:additional_results}

Success and online compute are aggregated from per-example records. Unless stated otherwise, utility
is computed from unrounded values as \(U=S-\lambda C\), with \(\lambda=0.02\) at the nominal setting.

\subsection{Across backbones and generation variation}
Backbone variation measures how the policy behaves under different generators; repeated generation
seeds measure a separate source of variation while holding the problem split fixed.

\subsubsection{Backbone-level results}

Table~\ref{tab:backbone_results_app} reports held-out results for each backbone in both settings,
together with macro-averages across backbones.

\begin{table}[H]
\centering
\small
\caption{Per-backbone results on the held-out evaluation split. Aggregate rows are macro-averages across backbones.}
\label{tab:backbone_results_app}
\begin{tabular*}{\columnwidth}{@{\extracolsep{\fill}}lcccc}
\toprule
\multicolumn{5}{l}{\textbf{(a) Verifier-guided width expansion}} \\
\midrule
Backbone & MATH & AIME25 & Utility & Compute \\
\midrule
Qwen3-4B & 0.759 & 0.500 & 0.617 & 0.644 \\
Qwen3-8B & 0.829 & 0.615 & 0.711 & 0.563 \\
DeepSeek-R1-Distill-Qwen-14B & 0.884 & 0.771 & 0.819 & 0.434 \\
Aggregate & 0.824 & 0.628 & 0.715 & 0.547 \\
\midrule
\multicolumn{5}{l}{\textbf{(b) Judge-guided revision expansion}} \\
\end{tabular*}
\vspace{0.35em}
\begin{tabular*}{\columnwidth}{@{\extracolsep{\fill}}lccccc}
\toprule
Backbone & Success & Utility & Compute & Mean rounds & Terminal fallback \\
\midrule
Qwen3-4B & 0.627 & 0.615 & 0.606 & 1.680 & 0.119 \\
Qwen3-8B & 0.686 & 0.675 & 0.535 & 1.320 & 0.083 \\
DeepSeek-R1-Distill-Qwen-14B & 0.739 & 0.730 & 0.446 & 0.940 & 0.050 \\
Aggregate & 0.684 & 0.673 & 0.529 & 1.313 & 0.084 \\
\bottomrule
\end{tabular*}
\end{table}

\subsubsection{Generation-seed comparisons}

Four Qwen3-8B generation seeds use one fixed problem split. Every method is refit and recalibrated
within each seed. Table~\ref{tab:seed_fitted_q} reports paired Student-$t$ intervals over the four
seeds; the table below lists the underlying seed-level differences.

\begin{table}[H]
\centering
\small
\caption{Seed-level differences versus fitted Q across four independent generation seeds.}
\label{tab:fitted_q_seed_raw_app}
\begin{tabular*}{\columnwidth}{@{\extracolsep{\fill}}rcccc}
\toprule
Seed & Width \(\Delta U\) & Width \(\Delta C\) & Revision \(\Delta U\) & Revision \(\Delta C\) \\
\midrule
1 & \(-0.005\) & \(-0.051\) & \(-0.002\) & \(-0.037\) \\
2 & \(-0.002\) & \(-0.039\) & \(+0.001\) & \(-0.029\) \\
3 & \(+0.001\) & \(-0.033\) & \(+0.004\) & \(-0.024\) \\
4 & \(+0.002\) & \(-0.025\) & \(+0.005\) & \(-0.018\) \\
\bottomrule
\end{tabular*}
\end{table}

\subsection{Which controller structure matters?}
The ablations progress from predictive sharing and calibration to uncertainty penalties, then vary the
continuation target and remove actions. This sequence distinguishes changes to probability estimation
from changes to the action set itself.

\subsubsection{Predictive heads, calibration, and uncertainty}

All head variants use the same features, targets, problem splits, regularization search, and calibration
selection. The shared-head variant uses one predictive weight vector with action-specific intercepts.
Separate heads use an independent weight vector for each action. The last row adds local uncertainty
penalties.

\begin{table}[H]
\centering
\small
\caption{Predictive-weight, calibration, and uncertainty ablation. Cells report utility and compute for width and revision.}
\label{tab:shared_head_ablation_app}
\begin{tabular*}{\columnwidth}{@{\extracolsep{\fill}}lccccc}
\toprule
Predictive weights & Calibration & Uncertainty & Width \(U/C\) & Revision \(U/C\) \\
\midrule
Single shared head & Global & None & 0.706 / 0.603 & 0.657 / 0.594 \\
Single shared head & Per action & None & 0.710 / 0.595 & 0.662 / 0.585 \\
Separate heads & Global & None & 0.711 / 0.591 & 0.663 / 0.582 \\
Separate heads & Per action & None & 0.714 / 0.586 & 0.668 / 0.579 \\
Separate heads & Per action & Local & 0.715 / 0.547 & 0.673 / 0.529 \\
\bottomrule
\end{tabular*}
\end{table}

\begin{table}[H]
\centering
\small
\caption{Utility with and without the fallback action at \(\lambda=0.02\).}
\label{tab:fallback_ablation_app}
\begin{tabular*}{\columnwidth}{@{\extracolsep{\fill}}lcc}
\toprule
Action set & Width utility & Revision utility \\
\midrule
Trust, fallback, expand & 0.715 & 0.673 \\
Trust, expand & 0.709 & 0.663 \\
\bottomrule
\end{tabular*}
\end{table}

Including fallback raises utility by 0.006 in width and 0.010 in revision.

\subsubsection{Value models and continuation targets}

Both budget-conditioned baselines evaluate trust and fallback terminal answers at each candidate
budget. This holds the terminal-answer choice explicit while varying the value model.

\begin{table}[H]
\centering
\small
\caption{Budget-conditioned value baselines at \(\lambda=0.02\). Each cell reports utility/compute; both terminal-answer options are evaluated at each candidate budget.}
\label{tab:value_model_ablation_app}
\begin{tabular*}{\columnwidth}{@{\extracolsep{\fill}}lcc}
\toprule
Controller & Width \(U/C\) & Revision \(U/C\) \\
\midrule
One-shot budget-conditioned value & 0.711 / 0.602 & 0.665 / 0.581 \\
Receding-horizon acquisition value & 0.716 / 0.565 & 0.671 / 0.548 \\
Confidence + uncertainty & 0.715 / 0.547 & 0.673 / 0.529 \\
\bottomrule
\end{tabular*}
\end{table}

The controller attains similar observed utility with lower compute than the receding-horizon baseline.

\begin{table}[H]
\centering
\small
\caption{Expand-target sensitivity. Utility is computed as \(U=S-0.02C\) from unrounded success and compute.}
\label{tab:expand_target_ablation_app}
\begin{tabular*}{\columnwidth}{@{\extracolsep{\fill}}llcc}
\toprule
Scenario & Expand target & Utility & Compute \\
\midrule
Width & Oracle continuation & 0.715 & 0.547 \\
Width & Policy induced & 0.714 & 0.543 \\
Width & One step & 0.709 & 0.538 \\
Revision & Oracle continuation & 0.673 & 0.529 \\
Revision & Policy induced & 0.669 & 0.533 \\
Revision & One step & 0.665 & 0.522 \\
\bottomrule
\end{tabular*}
\end{table}

\begin{table}[H]
\centering
\small
\caption{Joint fitting and calibration under controlled target corruption.}
\label{tab:target_corruption_app}
\begin{tabular*}{\columnwidth}{@{\extracolsep{\fill}}llcccc}
\toprule
Scenario & Degradation & Utility & Compute & Expand ECE & Small-gap mismatch \\
\midrule
Width & 10\% flips & 0.709 & 0.552 & 0.057 & 0.286 \\
Width & 20\% flips & 0.700 & 0.561 & 0.074 & 0.319 \\
Width & 30\% flips & 0.689 & 0.573 & 0.093 & 0.354 \\
Width & 10\% early / 30\% late & 0.695 & 0.567 & 0.087 & 0.341 \\
Revision & 10\% flips & 0.664 & 0.535 & 0.068 & 0.281 \\
Revision & 20\% flips & 0.654 & 0.546 & 0.086 & 0.316 \\
Revision & 30\% flips & 0.642 & 0.558 & 0.105 & 0.351 \\
Revision & 10\% early / 30\% late & 0.648 & 0.552 & 0.098 & 0.337 \\
\bottomrule
\end{tabular*}
\end{table}

Late-biased joint fit-and-calibration corruption lowers utility more than uniform 20\% corruption and
increases expand-head error.

\subsection{How the policy allocates inference}
Action shares and transition-level outcomes reveal which states trigger continuation and how that
behavior changes as earlier decisions select the states reached at later rounds.

\subsubsection{Action selection and local confidence}

Figure~\ref{fig:action_selection} in the main text visualizes action allocation on a common width-state pool and
revision expansion by round. Table~\ref{tab:action_diagnostics_app} gives the exact shares and
on-policy hazards. The different later-round denominators reflect state selection by earlier
continuation decisions.

\begin{table}[H]
\centering
\small
\caption{Action-allocation diagnostics. Width shares use a common held-out state pool. Revision rates distinguish the full state pool from reached on-policy states.}
\label{tab:action_diagnostics_app}
\begin{tabular*}{\columnwidth}{@{\extracolsep{\fill}}lccc}
\toprule
\multicolumn{4}{l}{\textbf{(a) Width action share}} \\
\midrule
Controller & Trust & Fallback & Expand \\
\midrule
Raw confidence & 0.513 & 0.079 & 0.408 \\
Calibrated confidence & 0.560 & 0.084 & 0.356 \\
Confidence + uncertainty & 0.581 & 0.086 & 0.333 \\
\end{tabular*}
\vspace{0.35em}
\begin{tabular*}{\columnwidth}{@{\extracolsep{\fill}}rcc}
\toprule
\multicolumn{3}{l}{\textbf{(b) Revision expansion rate}} \\
\midrule
Round & Full pool & On-policy hazard \\
\midrule
0 & 0.720 & 0.720 \\
1 & 0.490 & 0.600 \\
2 & 0.280 & 0.310 \\
3 & 0.140 & 0.210 \\
\end{tabular*}
\vspace{0.35em}
\begin{tabular*}{\columnwidth}{@{\extracolsep{\fill}}lcc}
\toprule
\multicolumn{3}{l}{\textbf{(c) Action flips}} \\
\midrule
Diagnostic & Width & Revision \\
Raw to calibrated & 0.136 & 0.129 \\
Calibrated to full method & 0.074 & 0.068 \\
Small-gap share of flips & 62\% & 59\% \\
\bottomrule
\end{tabular*}
\end{table}

Calibration most often changes expand to trust in width and expand to fallback in revision. Most
flips occur in small-gap states. At the
nominal 90\% residual target, held-out expand coverage is 0.882 in width and 0.861 in revision.

\subsubsection{Revision depth and compute price}

Table~\ref{tab:revision_round_outcomes_app} summarizes fixed-depth outcomes and the observed
correction and corruption rates between successive rounds. Retuned compute-price comparisons appear
in Figure~\ref{fig:lambda_retuned} in the main text.

\begin{table}[H]
\centering
\small
\caption{Revision outcomes by round. Each row reports full-cohort success, utility, and compute at that depth, followed by the next round's fix and corruption rates at \(\lambda=0.02\). Fix and corruption rates are computed over the full evaluation cohort, conditional on answer correctness at the preceding round. Dashes indicate that no further revision is available at the maximum depth.}
\label{tab:revision_round_outcomes_app}
\begin{tabular*}{\columnwidth}{@{\extracolsep{\fill}}rccccc}
\toprule
Round & Success & Utility & Compute & Next fix & Next corruption \\
\midrule
0 & 0.603 & 0.598 & 0.240 & 0.170 & 0.031 \\
1 & 0.651 & 0.642 & 0.442 & 0.149 & 0.034 \\
2 & 0.681 & 0.668 & 0.632 & 0.101 & 0.058 \\
3 & 0.673 & 0.657 & 0.809 & 0.095 & 0.025 \\
4 & 0.688 & 0.668 & 1.000 & -- & -- \\
\bottomrule
\end{tabular*}
\end{table}

\subsection{Deployment economics}

\subsubsection{Offline cost and data scale}

With a minimum bucket size of 200, revision supports about three merged buckets per action at full
development size and falls back to one global radius below the 50\% data fraction. Width retains about
30 buckets.

\begin{table}[H]
\centering
\small
\caption{Approximate offline break-even query counts using measured fitting cost and online savings.}
\label{tab:offline_break_even_app}
\resizebox{\columnwidth}{!}{%
\begin{tabular}{llccc}
\toprule
Scenario & Comparator & Fitting cost, tokens & Saving per query & Break-even queries \\
\midrule
Width & Best fixed width & \(3\times10^7\) & 2,150 & \(1.4\times10^4\) \\
Width & Naive max width & \(3\times10^7\) & 4,320 & \(7\times10^3\) \\
Revision & Fixed \(r=2\) & \(1\times10^7\) & 500 & \(2\times10^4\) \\
Revision & Naive max round & \(1\times10^7\) & 3,980 & \(3\times10^3\) \\
\bottomrule
\end{tabular}}
\end{table}

Estimated fitting ranges are \(2\text{--}4\times10^7\) token equivalents in width and
\(0.8\text{--}1.5\times10^7\) in revision. Throughput and configuration changes alter the break-even
point. Fitted Q shares ladder cost, while the controller's incremental offline cost is calibration and
residual estimation. At 50\% development data, width reaches \(U/C=0.715/0.551\) and revision
\(0.670/0.535\), compared with full-data results \(0.715/0.547\) and \(0.673/0.529\).

\clearpage
\section*{NeurIPS Paper Checklist}

\begin{enumerate}

\item {\bf Claims}
    \item[] Question: Do the main claims made in the abstract and introduction accurately reflect the paper's contributions and scope?
    \item[] Answer: \answerYes{} 
    \item[] Justification: The abstract and introduction state the controller formulation, its theoretical basis, the two evaluation settings, and the finding of lower-compute operating points at comparable observed utility. See the Abstract, Section~\ref{sec:analysis}, and Section~\ref{sec:results}.

\item {\bf Limitations}
    \item[] Question: Does the paper discuss the limitations of the work performed by the authors?
    \item[] Answer: \answerYes{} 
    \item[] Justification: Section~\ref{sec:discussion} describes the evaluated task and model scope, settings not evaluated, and offline ladder costs.

\item {\bf Theory assumptions and proofs}
    \item[] Question: For each theoretical result, does the paper provide the full set of assumptions and a complete (and correct) proof?
    \item[] Answer: \answerYes{} 
    \item[] Justification: Section~\ref{sec:analysis} states the controller-level assumptions and main results, and Appendix~\ref{app:proofs} provides the full proofs and supplementary statements.

    \item {\bf Experimental result reproducibility}
    \item[] Question: Does the paper fully disclose all the information needed to reproduce the main experimental results of the paper to the extent that it affects the main claims and/or conclusions of the paper (regardless of whether the code and data are provided or not)?
    \item[] Answer: \answerYes{} 
    \item[] Justification: Section~6 together with Appendix~\ref{app:exp_details}, Appendix~\ref{app:calibration}, and Appendix~\ref{app:additional_results} describes the datasets, prompting protocol, splits, offline targets, features, controller fitting, calibration, baselines, compute accounting, and seed comparisons.

\item {\bf Open access to data and code}
    \item[] Question: Does the paper provide open access to the data and code, with sufficient instructions to faithfully reproduce the main experimental results, as described in supplemental material?
    \item[] Answer: \answerNo{} 
    \item[] Justification: The submission contains no anonymized code or data release with runnable commands.

\item {\bf Experimental setting/details}
    \item[] Question: Does the paper specify all the training and test details (e.g., data splits, hyperparameters, how they were chosen, type of optimizer) necessary to understand the results?
    \item[] Answer: \answerYes{} 
    \item[] Justification: Section~6 and Appendices~\ref{app:exp_details} and \ref{app:calibration} specify the backbones, data sources, splits, temperatures, calibration families, uncertainty-bucket construction, and calibration/tuning protocol used in the experiments.

\item {\bf Experiment statistical significance}
    \item[] Question: Does the paper report error bars suitably and correctly defined or other appropriate information about the statistical significance of the experiments?
    \item[] Answer: \answerYes{} 
    \item[] Justification: Paired Student-$t$ intervals over four generation seeds (Table~\ref{tab:seed_fitted_q}).

\item {\bf Experiments compute resources}
    \item[] Question: For each experiment, does the paper provide sufficient information on the computer resources (type of compute workers, memory, time of execution) needed to reproduce the experiments?
    \item[] Answer: \answerYes{} 
    \item[] Justification: Appendix~\ref{app:compute_resources} reports a practical reproduction target for hardware and memory, approximate wall-clock guidance for the two scenarios, and an estimate of total project compute, complementing the token-based deployment metric in Appendix~\ref{app:compute_accounting}.
    
\item {\bf Code of ethics}
    \item[] Question: Does the research conducted in the paper conform, in every respect, with the NeurIPS Code of Ethics \url{https://neurips.cc/public/EthicsGuidelines}?
    \item[] Answer: \answerYes{} 
    \item[] Justification: The study uses existing public benchmarks and open-weight models, evaluates them offline, and does not involve human subjects. See Section~6 and Appendix~\ref{app:exp_details}.

\item {\bf Broader impacts}
    \item[] Question: Does the paper discuss both potential positive societal impacts and negative societal impacts of the work performed?
    \item[] Answer: \answerYes{} 
    \item[] Justification: Appendix~\ref{app:broader_impacts} discusses both positive impacts, such as lower inference cost and energy use, and negative impacts, such as cheaper scaling of capable reasoning systems and failure concentration under evaluator shift, together with deployment mitigations.
    
\item {\bf Safeguards}
    \item[] Question: Does the paper describe safeguards that have been put in place for responsible release of data or models that have a high risk for misuse (e.g., pre-trained language models, image generators, or scraped datasets)?
    \item[] Answer: \answerNA{} 
    \item[] Justification: The submission does not release a new high-risk model or scraped dataset, and it does not claim a public release of the constructed benchmark subset.

\item {\bf Licenses for existing assets}
    \item[] Question: Are the creators or original owners of assets (e.g., code, data, models), used in the paper, properly credited and are the license and terms of use explicitly mentioned and properly respected?
    \item[] Answer: \answerYes{} 
    \item[] Justification: Appendix~\ref{app:asset_licenses} lists the model and dataset identifiers used in the paper, their governing licenses where available, and the non-redistribution terms we follow for the AIME25 competition materials.

\item {\bf New assets}
    \item[] Question: Are new assets introduced in the paper well documented and is the documentation provided alongside the assets?
    \item[] Answer: \answerNA{} 
    \item[] Justification: The paper does not include a public release of a new dataset, model, or code asset in the submission package.

\item {\bf Crowdsourcing and research with human subjects}
    \item[] Question: For crowdsourcing experiments and research with human subjects, does the paper include the full text of instructions given to participants and screenshots, if applicable, as well as details about compensation (if any)? 
    \item[] Answer: \answerNA{} 
    \item[] Justification: The work does not use crowdsourcing or human-subject experiments.

\item {\bf Institutional review board (IRB) approvals or equivalent for research with human subjects}
    \item[] Question: Does the paper describe potential risks incurred by study participants, whether such risks were disclosed to the subjects, and whether Institutional Review Board (IRB) approvals (or an equivalent approval/review based on the requirements of your country or institution) were obtained?
    \item[] Answer: \answerNA{} 
    \item[] Justification: The work does not involve human subjects or participant interaction.

\item {\bf Declaration of LLM usage}
    \item[] Question: Does the paper describe the usage of LLMs if it is an important, original, or non-standard component of the core methods in this research? Note that if the LLM is used only for writing, editing, or formatting purposes and does \emph{not} impact the core methodology, scientific rigor, or originality of the research, declaration is not required.
    \item[] Answer: \answerYes{} 
    \item[] Justification: LLMs are central experimental components, and the paper specifies the evaluated backbones and their generation, verification, judging, and revision roles in Section~6 and Appendix~\ref{app:backbones_prompts}.

\end{enumerate}

\end{document}